\documentclass[9pt,onecolumn]{IEEEtran}

\ifCLASSOPTIONcompsoc
 \usepackage[nocompress]{cite}
\else
\fi

\usepackage[pdftex]{graphicx}
\graphicspath{{./Fig/}{./Figs/}{./NewFigs/}{./AppendFigs/NewFigs/}{./FinalFigures/}{./Author_Photo/}}

\usepackage{graphicx}
\usepackage{amsmath}
\usepackage{amssymb}

\usepackage{mathrsfs}
\usepackage{amsfonts}
\usepackage{amsthm}

\usepackage{bm}
\usepackage[section]{placeins}

\usepackage[ruled,vlined]{algorithm2e}

\let\savedalgorithm\algorithm
\let\savedendalgorithm\endalgorithm

\usepackage{setspace}

\newcommand{\comment}[1]{}

\newcommand{\ninesevenhao}{\fontsize{8pt}{8pt}\selectfont}

\renewcommand{\vspace}[1]{}

\begin{document}

\begin{spacing}{1}

\title{Incremental Learning of 3D-DCT Compact Representations for Robust Visual Tracking}

\author{ 
\ninesevenhao
{Xi Li$^{\dag}$, Anthony Dick$^{\dag}$, Chunhua Shen$^{\dag}$, Anton van den Hengel$^{\dag}$, Hanzi Wang$^{\circ}$
\\
$^{\dag}$Australian Center for Visual Technologies, and School of Computer Sciences, University of Adelaide, Australia\\
$^{\circ}$Center for Pattern Analysis and Machine Intelligence, and Fujian Key Laboratory of the Brain-like Intelligent Systems, Xiamen University, China\\
}
}

\markboth{IEEE Transactions on Pattern Analysis and Machine Intelligence}
{Manuscript}

\IEEEcompsoctitleabstractindextext{
\begin{abstract}

Visual tracking usually requires an object appearance model that is robust to
changing illumination, pose and other factors encountered in video.
Many recent trackers utilize appearance samples in previous frames
to form the bases upon which the object appearance model is built.
This approach has the following limitations:
(a) the bases are data driven, so they can be easily corrupted;
and (b) it is difficult to robustly update the bases in challenging situations.

In this paper, we construct an appearance model using the 3D discrete cosine transform (3D-DCT). The 3D-DCT is based on a set of cosine basis functions,
which are determined by the dimensions of the 3D signal
and thus independent of the input video data.
In addition, the 3D-DCT can
generate a compact energy spectrum whose high-frequency coefficients are sparse if the appearance samples are similar.
By
discarding these high-frequency coefficients, we simultaneously obtain a compact 3D-DCT based object representation
and a signal reconstruction-based similarity measure (reflecting the information loss from signal reconstruction).
To efficiently update the object representation, we propose an incremental 3D-DCT algorithm, which
decomposes the 3D-DCT into successive operations of the 2D discrete cosine transform (2D-DCT)
and 1D discrete cosine transform (1D-DCT) on the input video
data. As a result, the incremental 3D-DCT algorithm only needs to compute the 2D-DCT for
newly added frames as well as the 1D-DCT along the third dimension, which significantly reduces the computational complexity.
Based on this
incremental 3D-DCT algorithm, we design a discriminative criterion to evaluate the likelihood
of a test sample belonging to the foreground object. We then embed the discriminative criterion
into a particle filtering framework for object state inference over time.
Experimental results demonstrate the effectiveness and robustness of the proposed tracker.

\end{abstract}

}

\maketitle
\thispagestyle{empty}

\IEEEdisplaynotcompsoctitleabstractindextext
\IEEEpeerreviewmaketitle

\setcounter{page}{1}

\section{Introduction}
\label{sec:paper_introduction}

Visual tracking of a moving object is a fundamental problem in computer vision. It has a wide range of
applications including visual surveillance, human behavior analysis,
motion event detection, and video retrieval. Despite much effort on this topic, it remains a challenging problem because of object appearance variations due to illumination changes, occlusions, pose changes, cluttered and moving backgrounds, etc.
Thus, a crucial element of visual tracking is to use an effective object appearance model that is robust to such challenges.

Since it is difficult to explicitly model complex appearance changes,
a popular approach is to learn a low-dimensional subspace (e.g., eigenspace~\cite{Limy-Ross17,lixi-cvpr2008}),
which accommodates the
object's observed appearance variations.
This allows the appearance model to reflect the time-varying properties of object appearance during tracking (e.g.,
learning the appearance of the object from multiple observed poses).
By computing the sample-to-subspace distance (e.g., reconstruction
error~\cite{Limy-Ross17,lixi-cvpr2008}),
the approach can measure the information loss that results from projecting
a test sample to the low-dimensional subspace.
Using  the information loss, the approach can evaluate the likelihood of a test sample
belonging to the foreground object.
Since the approach is data driven,  it needs to compute
the subspace basis vectors as well as the corresponding coefficients.

Inspired by the success of subspace learning for visual tracking,
we propose an alternative object representation based
on the 3D discrete cosine transform (3D-DCT), which has
a set of fixed projection bases (i.e., cosine basis functions).
Using these fixed projection bases, the proposed object representation only needs to
compute the corresponding projection coefficients (3D-DCT
coefficients).  Compared with incremental principal component analysis~\cite{Limy-Ross17}, this leads to a much simpler
computational process, which is more robust to many types of appearance change and enables fast implementation.

\begin{figure}[t]
\vspace{-0.05cm}
\begin{center}
   \includegraphics[width=0.8\linewidth]{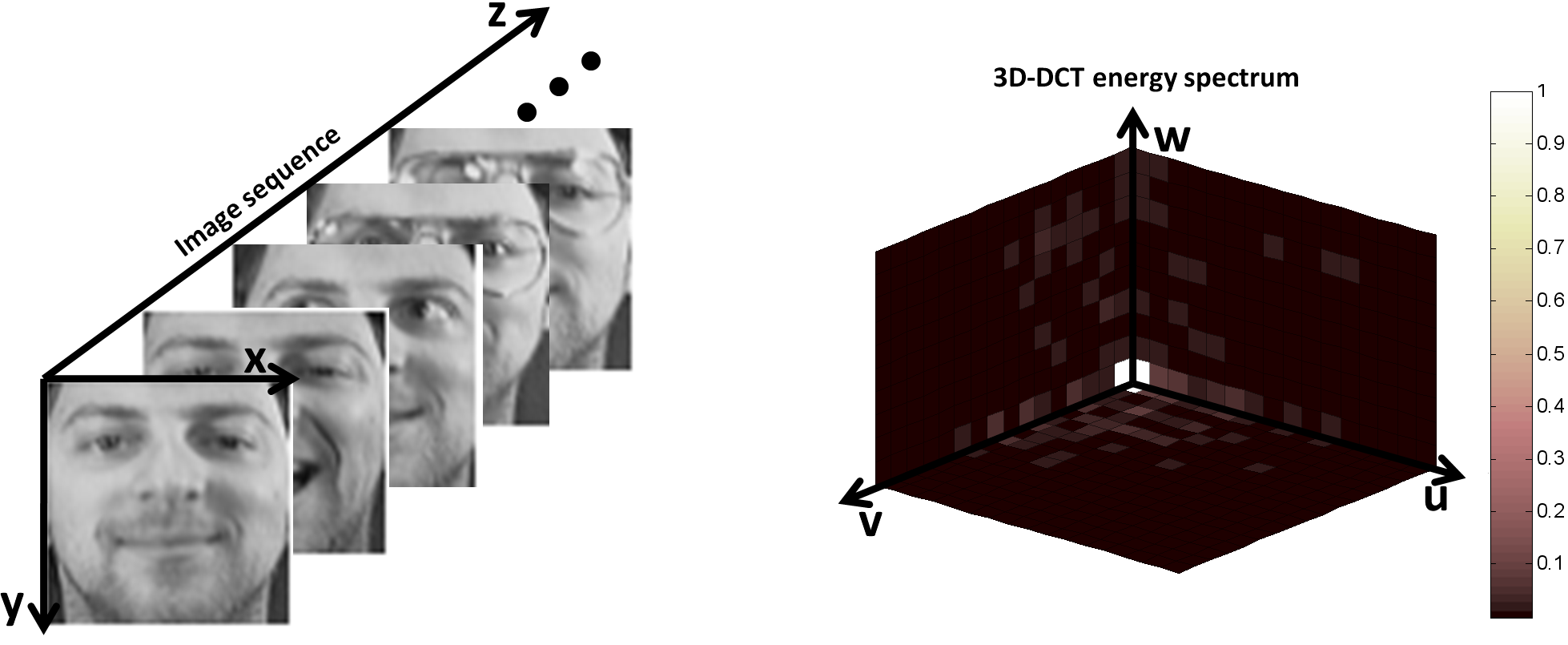}
\end{center}
\vspace{-0.8cm}
   \caption{Illustration of 3D-DCT's compactness. The left part shows a face image sequence, and the right part displays
   the corresponding energy spectrum of 3D-DCT. Clearly, it is seen from the right part that the energy spectrums of 3D-DCT are compact.}
    \label{fig:compact_energy_map} \vspace{-0.8cm}
\end{figure}

The DCT has a long history in the signal processing community as a tool for encoding images and video.
It has been shown to have desirable properties for representing video, many of which also make it a promising object representation for visual tracking in video:
\begin{itemize}
\item As illustrated in Fig.~\ref{fig:compact_energy_map}, the DCT leads to a compact object representation with
      sparse transform coefficients if a signal is self-correlated in both spatial and temporal dimensions.  This means that the reconstruction error induced by removing a subset of coefficients is typically small.
      Additionally, high-frequency image noise or rapid appearance changes are often isolated in a small number of coefficients;
\item The DCT's cosine basis functions are determined by the signal dimensions that are fixed at initialization.
      Thus, the DCT's cosine basis functions are fixed throughout tracking, resulting in a simple procedure of constructing
      the DCT-based object representation;
\item The DCT only requires single-level cosine decomposition to approximate the original signal, which again is computationally efficient and also lends itself to incremental calculation, which is useful for tracking.
\end{itemize}

Our idea is simply to represent a new sample by concatenating it with a collection of previous samples to form a 3D signal, and
calculating its coefficients in the 3D-DCT space with some high-frequency components removed.
Since the 3D-DCT encodes the temporal redundancy information of the 3D signal, the representation
can capture the correlation
between the new sample and the previous samples.
Given a compression ratio (derived from discarding some high-frequency components),
if the new sample can still be effectively reconstructed
with a relatively low reconstruction error, then it
is correlated with the previous samples and
is likely to be an object sample.
The fact that every sample is represented by using the same cosine basis functions makes it very easy to
perform the likelihood evaluations of samples.

The DCT is not the only choice for compact representations using
data-independent bases; others include
Fourier and wavelet basis functions, which are also widely used in signal processing.
The coefficients of these basis functions are capable of capturing the energy information at different frequencies.
For example, both sine and cosine basis functions are adopted by the discrete Fourier transform (DFT)
to generate the amplitude and phase frequency spectrums;
wavelet basis functions (e.g., Haar and Gabor) aim to capture
local detailed information (e.g., texture) of a signal at multiple resolutions by the wavelet transform (WT).
Although we do not conduct experiments with these functions in this work, they can be used in our framework with only minor modification.

Using the 3D-DCT object representation, we propose a discriminative learning based tracker. The main contributions
of this tracker are three-fold:
\begin{enumerate}
\item
We utilize the signal compression power of the 3D-DCT to construct a novel representation of a tracked object.
The representation retains the dense low-frequency 3D-DCT coefficients, and
discards the relatively sparse high-frequency 3D-DCT coefficients.
Based on this compact representation, the signal reconstruction error (measuring the information loss from signal reconstruction)
is used to evaluate the
likelihood of a test sample belonging to the foreground object given a set of training samples.
\item We propose an incremental 3D-DCT algorithm for efficiently updating the representation.
The incremental algorithm decomposes 3D-DCT into the successive operations of the 2D-DCT and 1D-DCT
on the input video data, and it only needs to
compute the 2D-DCT for newly added frames (referred to in Equ.~\eqref{eq:D_update}) as well as the 1D-DCT along the third dimension, resulting in high computational efficiency.
In particular, the cosine basis functions can be computed in advance, which significantly reduces the computational cost of the 3D-DCT.
\item
We design a discriminative criterion (referred to in Equ.~\eqref{eq:final_likelihood}) for predicting the confidence score of a test sample belonging to the foreground
object. The discriminative criterion considers both the foreground and the background 3D-DCT reconstruction likelihoods,
which enables the tracker to
capture useful discriminative information for
adapting to complicated appearance changes.
\end{enumerate}

\section{Related work}
Since our work focuses on learning compact object representations based on
the 3D-DCT,
we first discuss the DCT and its applications in
relevant research fields. Then, we briefly
review the related tracking algorithms using different types of
object representations.
As claimed in~\cite{AKJ-DCT,khayam-tr2003}, the DCT aims to use a set of mutually uncorrelated cosine basis functions to express a discrete signal
in a linear manner.
It has a wide range of applications in computer vision, pattern recognition, and
multimedia,
such
as face recognition~\cite{HAFED-LEVINE-IJCV2001},
image retrieval~\cite{Feng-PR2003,He-ICIP2009},
video object segmentation~\cite{Chen-Liu-Sun-Yang-TMM2008}, video caption localization~\cite{Zhong-Zhang-Jain-TPAMI2000}, etc.
In these applications, the DCT is typically
used for feature extraction, and aims to construct a compact DCT coefficient-based image representation that is
robust to complicated factors (e.g., facial geometry and illumination changes).
In this paper, we focus on how to construct an effective DCT-based object representation for
robust visual tracking.

In the field of visual tracking, researchers have  designed a variety of object representations,
which can be roughly classified into two categories: generative object representations
and discriminative object representations.

Recently, much work has been done in
constructing  generative object representations, including
the integral histogram~\cite{Adam-Fragment-2006}, kernel density
estimation~\cite{Shen-Kim-Wang-TCSVT2010},
mixture models~\cite{Wang-Suter-Schindler-PAMI2007, Jepson-Fleet-Yacoob5}, subspace
learning~\cite{Limy-Ross17,li2007robust},
linear representation~\cite{Meo-Ling-ICCV09,Liu-Yang-Huang-Meer-Gong-Kulikowski-eccv2010,Liu-Huang-Kulikowski-Yang-cvpr2011, Li-Shen-Shi-cvpr2011,licvpr2012}, visual
tracking decomposition~\cite{Kwon-Lee-CVPR2010}, covariance tracking~\cite{Porikli-Tuzel-Meer-CVPR2006,lixi-cvpr2008,Wu-Cheng-Wang-Lu-iccv2009}, and so on.
Some representative tracking algorithms based on generative object representations are reviewed as follows.
Jepson \emph{et al.}~\cite{Jepson-Fleet-Yacoob5} design a
more elaborate mixture model with an online EM algorithm to
explicitly model appearance changes during tracking.
Wang \emph{et al.}~\cite{Wang-Suter-Schindler-PAMI2007} present an
adaptive appearance model based on the Gaussian mixture model
in a joint spatial-color space.
Comaniciu \emph{et al.}~\cite{Comaniciu-Ramesh-Meer-TPAMI}  propose a kernel-based tracking algorithm
using the mean shift-based mode seeking procedure.
Following the work of \cite{Comaniciu-Ramesh-Meer-TPAMI}, some variants of the kernel-based tracking algorithm
are proposed, e.g., \cite{ Shen-Kim-Wang-TCSVT2010, Shen-Brooks-van-den-Hengel-TIP2007, Qu-Schonfeld-TIP2008}.
Ross \emph{et al.}~\cite{Limy-Ross17} propose a
generalized tracking framework based on the incremental
PCA (principal component analysis) subspace learning method with a sample mean update.
A sparse approximation based
tracking algorithm using $\ell_{1}$-regularized minimization is proposed
by Mei and Ling~\cite{Meo-Ling-ICCV09}. To achieve a real-time performance, Li \emph{et al.}~\cite{Li-Shen-Shi-cvpr2011}
present a compressive sensing $\ell_{1}$ tracker using an orthogonal matching
pursuit algorithm, which is up to 6000 times faster than~\cite{Meo-Ling-ICCV09}.

In contrast, another type of tracking algorithms try to
construct a variety of discriminative object representations, which
aim to maximize the inter-class
separability between the object and non-object regions using  discriminative
learning techniques, including SVMs~\cite{Avidan-2004, Tian-Zhang-Liu-ACCV2007,
Tang-Brennan-Tao-ICCV2007,li2011graph}, boosting~\cite{Grabner-Grabner-Bischof-BMVC2006,
Grabner-Grabner-Bischof-ECCV2008}, discriminative feature selection~\cite{Collins-Liu-Leordeanu-PAMI2005}, random
forest~\cite{Santner-Leistner-Saffari-Pock-Bischof-cvpr2010}, multiple instance
learning~\cite{Babenko-Yang-Belongie-cvpr2009}, spatial attention
learning~\cite{Fan-Wu-Dai-ECCV2010}, discriminative metric learning~\cite{Wang-Hua-Han-eccv2010,Jiang-Liu-Wu-TIP2011},
data-driven adaptation \cite{Yang-Fan-Fan-Wu-TIP2009}, etc.
Some popular tracking algorithms based on discriminative object representations
are described as follows.
Grabner \emph{et al.}~\cite{Grabner-Grabner-Bischof-BMVC2006} design an online AdaBoost classifier for
discriminative feature selection during tracking, resulting in the robustness to
the appearance variations caused by out-of-plane rotations and illumination changes.
To alleviate the model drifting problem with~\cite{Grabner-Grabner-Bischof-BMVC2006},
Grabner \emph{et al.}~\cite{Grabner-Grabner-Bischof-ECCV2008}
present a semi-supervised online boosting algorithm
for tracking.
Liu and Yu~\cite{Liu-Yu-ICCV2007}
present a gradient-based feature selection mechanism for online boosting learning,
leading to the higher tracking efficiency.
Avidan~\cite{Avidan-2007}
builds an
ensemble of online learned weak classifiers for pixel-wise classification, and then
employ mean shift for object localization.
Instead of using single-instance boosting,
Babenko \emph{et al.}~\cite{Babenko-Yang-Belongie-cvpr2009} present a tracking system
based on online multiple instance boosting, where an object is represented as
a set of image patches.
Besides, SVM-based object representations have also attracted much attention in recent years.
Based on off-line SVM learning, Avidan \cite{Avidan-2004} proposes a tracking algorithm
for distinguishing a target vehicle from
backgrounds.
Later, Tian \emph{et al.}~\cite{Tian-Zhang-Liu-ACCV2007} present a tracking system based on
an ensemble of linear SVM classifiers, which can be adaptively weighted according to their discriminative
abilities during different periods.
Instead of using supervised learning, Tang \emph{et al.}~\cite{Tang-Brennan-Tao-ICCV2007} present an online semi-supervised learning
based tracker, which constructs two feature-specific SVM classifiers in a
co-training framework.

As our tracking algorithm is based on the DCT,  we give a brief review of the discrete cosine transform and its three basic versions
for 1D, 2D, and 3D signals in the next section.

\section{The 3D-DCT for object representation}
\label{sec:3ddct}

We first give an introduction to the 3D-DCT in Section~\ref{sec:intro_dct}.
Then,
we derive and formulate the DCT's matrix forms (used for object representation) in Section~\ref{sec:3d_dct_reformulation}.
Next, we address the problem of how to use the 3D-DCT as a compact object representation in Section~\ref{sec:compact_object_representation}.
Finally, we propose an incremental 3D-DCT algorithm to efficiently compute the 3D-DCT in Section~\ref{sec:incremental_3D_dct}.

\subsection{3D-DCT definitions and notations}
\label{sec:intro_dct}

The goal of the discrete cosine transform (DCT) is to express a discrete signal, such as a digital image or video, as a linear combination of
mutually uncorrelated cosine basis functions (CBFs), each of which encodes frequency-specific information of the discrete signal.

We briefly define the 1D-DCT, 2D-DCT, and 3D-DCT, which are applied to 1D signal $\left(f_{\mathbf{I}}(x)\right)_{x=0}^{N_{1}-1}$,
2D signal $\left(f_{\mathbf{II}}(x, y)\right)_{N_{1}\times N_{2}}$
and 3D signal
$\left(f_{\mathbf{III}}(x, y, z)\right)_{N_{1}\times N_{2} \times N_{3}}$
respectively:

\begin{equation}
C_{\mathbf{I}}(u) = \alpha_{1}(u)\sum_{x=0}^{N_{1}-1}f_{\mathbf{I}}(x)\cos\left[\frac{\pi(2x+1)u}{2N_{1}}\right],
\label{eq:dct_coef} \vspace{-0.15cm}
\end{equation}
\begin{equation}
C_{\mathbf{II}}(u, v) =  \alpha_{1}(u)\alpha_{2}(v)\sum_{x=0}^{N_{1}-1}\sum_{y=0}^{N_{2}-1}f_{\mathbf{II}}(x, y)
\cos\left[\frac{\pi(2x+1)u}{2N_{1}}\right]\cos\left[\frac{\pi(2y+1)v}{2N_{2}}\right],
\label{eq:2D_DCT} \vspace{-0.15cm}
\end{equation}
\begin{equation}
\begin{array}{l}
C_{\mathbf{III}}(u, v, w) = \alpha_{1}(u)\alpha_{2}(v)\alpha_{3}(w)\sum_{x=0}^{N_{1}-1}\sum_{y=0}^{N_{2}-1}\sum_{z=0}^{N_{3}-1}f_{\mathbf{III}}(x, y, z)\\
\hspace{3.1cm}\cdot\left\{
\cos\left[\frac{\pi(2x+1)u}{2N_{1}}\right]\cos\left[\frac{\pi(2y+1)v}{2N_{2}}\right]\cos\left[\frac{\pi(2z+1)w}{2N_{3}}\right]\right\},
\end{array}
\label{eq:3D_DCT} \vspace{-0.15cm}
\end{equation}
where  $u\in \{0, 1,  \ldots, N_{1}-1\}$, $v\in \{0, 1,  \ldots, N_{2}-1\}$,
$w\in \{0, 1,  \ldots, N_{3}-1\}$
and $\alpha_{k}(u)$ is defined as
\begin{equation}
\alpha_{k}(u) = \left\{
\begin{array}{ll}
\sqrt{\frac{1}{N_{k}}}, & \mbox{if} \thickspace u=0;\\
\sqrt{\frac{2}{N_{k}}}, & \mbox{otherwise};
\end{array}
\right.
\label{eq:alpha} \vspace{-0.15cm}
\end{equation}
where $k$ is a positive integer.

The corresponding inverse DCTs (referred to as 1D-IDCT, 2D-IDCT, and 3D-IDCT) are defined as:

\begin{equation}
f_{\mathbf{I}}(x) = \sum_{u=0}^{N_{1}-1}C_{\mathbf{I}}(u)\underbrace{\alpha_{1}(u)\cos\left[\frac{\pi(2x+1)u}{2N_{1}}\right]}_{\mbox{1D-DCT CBF}},
\vspace{-0.15cm}
\label{eq:inverse_1D_DCT}
\end{equation}
\begin{equation}
f_{\mathbf{II}}(x, y) = \sum_{u=0}^{N_{1}-1}\sum_{v=0}^{N_{2}-1}C_{\mathbf{II}}(u, v)\underbrace{\alpha_{1}(u)\alpha_{2}(v)
\cos\left[\frac{\pi(2x+1)u}{2N_{1}}\right]\cos\left[\frac{\pi(2y+1)v}{2N_{2}}\right]}_{\mbox{2D-DCT CBF}}, \vspace{-0.15cm}
\label{eq:inverse_2D_DCT}
\end{equation}
\begin{equation}
\begin{array}{l}
\hspace{-1.3cm}
f_{\mathbf{III}}(x, y, z) = \sum_{w=0}^{N_{3}-1}\sum_{u=0}^{N_{1}-1}\sum_{v=0}^{N_{2}-1}C_{\mathbf{III}}(u, v, w)\cdot\\
\hspace{-0.6cm}
\underbrace{\alpha_{1}(u)\alpha_{2}(v)\alpha_{3}(w)\cos\left[\frac{\pi(2x+1)u}{2N_{1}}\right]\cos\left[\frac{\pi(2y+1)v}{2N_{2}}\right]\cos\left[\frac{\pi(2z+1)w}{2N_{3}}\right]}_{\mbox{3D-DCT CBF}}.
\end{array}
\hspace{-0.6cm}
\label{eq:I3D_DCT} \vspace{-0.25cm}
\end{equation}

The low-frequency CBFs reflect the larger-scale energy information (e.g., mean value) of the discrete signal, while
the high-frequency CBFs capture the smaller-scale energy information (e.g., texture) of the discrete signal.
Based on these CBFs, the original discrete signal can be transformed into a DCT coefficient space whose dimensions
are mutually uncorrelated.
Furthermore, the output of the DCT is typically sparse, which is useful for signal compression and also for tracking, as will be shown in the following sections.

\subsection{3D-DCT matrix formulation \label{sec:DCT_matrx_reformulation}}
\label{sec:3d_dct_reformulation}

Let $\mathbf{C}_{\mathbf{I}} = \left(C_{\mathbf{I}}(0), C_{\mathbf{I}}(1), \ldots, C_{\mathbf{I}}(N_{1}-1)\right)^{T}$ denote the 1D-DCT
coefficient column vector.
Based on Equ.~\eqref{eq:dct_coef},  $\mathbf{C}_{\mathbf{I}}$ can be rewritten in a matrix form:
$\mathbf{C}_{\mathbf{I}} = \mathbf{A}_{1}\mathbf{f}$, where $\mathbf{f}$ is a column vector: $\mathbf{f}=(f_{\mathbf{I}}(0), f_{\mathbf{I}}(1), \ldots, f_{\mathbf{I}}(N_{1}-1))^{T}$
and $\mathbf{A}_{1}=\left(a_{1}(u, x)\right)_{N_{1}\times N_{1}}$ is a cosine basis matrix whose entries are given by: \vspace{-0.1cm}
\begin{equation}
a_{1}(u, x) = \alpha_{1}(u)\cos\left[\frac{\pi(2x+1)u}{2N_{1}}\right].
\label{eq:A} \vspace{-0.1cm}
\end{equation}
The matrix form of 1D-IDCT can be written as:
$\mathbf{f} = \mathbf{A}_{1}^{-1}\mathbf{C}_{\mathbf{I}}$.
Since $\mathbf{A}_{1}$ is an orthonormal matrix, $\mathbf{f} = \mathbf{A}_{1}^{T}\mathbf{C}_{\mathbf{I}}$.

The 2D-DCT coefficient matrix
$\mathbf{C}_{\mathbf{II}}=(C_{\mathbf{II}}(u,v))_{N_{1}\times N_{2}}$
corresponding to Equ.~\eqref{eq:2D_DCT} is formulated as:
$\mathbf{C}_{\mathbf{II}} = \mathbf{A}_{1}\mathbf{F}\mathbf{A}_{2}^{T}$, where $\mathbf{F} = (f_{\mathbf{II}}(x,y))_{N_{1} \times N_{2}}$  is the original 2D signal,
$\mathbf{A}_{1}$ is defined in Equ.~\eqref{eq:A}, and $\mathbf{A}_{2}$ is defined as $(a_{2}(v, y))_{N_{2}\times N_{2}}$ such that \vspace{-0.1cm}
\begin{equation}
a_{2}(v, y) = \alpha_{2}(v)\cos\left[\frac{\pi(2y+1)v}{2N_{2}}\right].
\label{eq:A2} \vspace{-0.1cm}
\end{equation}
The matrix form of the 2D-IDCT can be expressed as:
$\mathbf{F} = \mathbf{A}_{1}^{-1}\mathbf{C}_{\mathbf{II}}(\mathbf{A}_{2}^{T})^{-1}$.
Since the DCT basis functions are orthonormal, we have
$\mathbf{F} = \mathbf{A}_{1}^{T}\mathbf{C}_{\mathbf{II}}\mathbf{A}_{2}$.

Similarly, the 3D-DCT can be decomposed into a succession of the 2D-DCT and 1D-DCT
operations.
Let $\mathcal{F}= (f_{\mathbf{III}}(x,y,z))_{N_{1} \times N_{2}\times N_{3}}$ denote a 3D signal.
Mathematically, $\mathcal{F}$ can be viewed as a three-order tensor, i.e., $\mathcal{F}\in \mathcal{R}^{N_{1}\times N_{2} \times N_{3}}$.
Consequently, we need to introduce terminology for the mode-$m$ product defined in tensor algebra~\cite{Levy-Lindenbaum-JMAA2000}.
Let $\mathcal{B}\in \mathcal{R}^{I_{1}\times I_{2} \times \ldots \times I_{M}}$ denote an $M$-order tensor,
each element of which is represented as $b(i_{1},\ldots, i_{m}\ldots, i_{M})$ with $1 \leq i_{m}\leq I_{m}$. In tensor terminology,
each dimension of a tensor is associated with a ``\emph{mode}''.
The mode-$m$ product
of the tensor $\mathcal{B}$ by a matrix $\mathbf{\Phi}=(\phi(j_{m}, i_{m}))_{J_{m}\times I_{m}}$ is denoted as
$\mathcal{B} \times_{m}\mathbf{\Phi}$ whose entries are as follows:
\vspace{-0.1cm}
\begin{equation}
\left(\mathcal{B}\times_{m} \mathbf{\Phi}\right)(i_{1}, \ldots, i_{m-1}, j_{m}, i_{m+1}, \ldots, i_{M})= \sum_{i_{m}}b(i_{1}, \ldots, i_{m}, \ldots, i_{M}) \phi(j_{m},i_{m}),
\vspace{-0.1cm}
\end{equation}
where
$\times_{m}$ is the mode-$m$ product operator and
$1\leq m \leq M$.
Given two matrices $\mathbf{G}\in \mathcal
{R}^{J_{m}\times I_{m}}$ and $\mathbf{H}\in \mathcal {R}^{J_{n}\times I_{n}}$ such that $m\neq n$,
the following relation holds: \vspace{-0.1cm}
\begin{equation}
(\mathcal{B} \times_{m} \mathbf{G}) \times_{n} \mathbf{H}
       =(\mathcal{B} \times_{n} \mathbf{H}) \times_{m} \mathbf{G}=\mathcal{B} \times_{m} \mathbf{G} \times_{n} \mathbf{H}. \vspace{-0.1cm}
\end{equation}
Based on the above tensor algebra, the 3D-DCT coefficient matrix $\mathbf{C}_{\mathbf{III}}=(C_{\mathbf{III}}(u,v,w))_{N_{1}\times N_{2}\times N_{3}}$ can be formulated as:
$\mathbf{C}_{\mathbf{III}} = \mathcal{F}\times_{1}\mathbf{A}_{1}\times_{2}\mathbf{A}_{2}\times_{3}\mathbf{A}_{3}$, where $\mathbf{A}_{3}=(a_{3}(w, z))_{N_{3}\times N_{3}}$ has a similar definition to $\mathbf{A}_{1}$ and $\mathbf{A}_{2}$: \vspace{-0.1cm}
\begin{equation}
a_{3}(w, z) = \alpha_{3}(w)\cos\left[\frac{\pi(2z+1)w}{2N_{3}}\right].
\vspace{-0.1cm}
\label{eq:A3a}
\end{equation}
Accordingly, 3D-IDCT is formulated as: $\mathcal{F} = \mathbf{C}_{\mathbf{III}}\times_{1}\mathbf{A}_{1}^{-1}\times_{2}\mathbf{A}^{-1}_{2}\times_{3}\mathbf{A}^{-1}_{3}$.
Since $\mathbf{A}_{k} (1\leq k \leq 3)$ is an orthonormal matrix,  $\mathcal{F}$ can be rewritten as: \vspace{-0.1cm}
\begin{equation}
\mathcal{F} = \mathbf{C}_{\mathbf{III}}\times_{1}\mathbf{A}_{1}^{T}\times_{2}\mathbf{A}^{T}_{2}\times_{3}\mathbf{A}^{T}_{3}.
\label{eq:F_tensor_recons} \vspace{-0.1cm}
\end{equation}
In fact, the 1D-DCT and 2D-DCT are two special cases of the 3D-DCT because 1D vectors and 2D matrices are 1-order and 2-order tensors, respectively,
namely, $\mathbf{f}\times_{1}\mathbf{A}_{1} = \mathbf{A}_{1}\mathbf{f}$ and $\mathbf{F}\times_{1}\mathbf{A}_{1}\times_{2}\mathbf{A}_{2} = \mathbf{A}_{1}\mathbf{F}\mathbf{A}_{2}^{T}$.

\subsection{Compact object representation using the 3D-DCT }
\label{sec:compact_object_representation}

For visual tracking, an input video sequence can be viewed as 3D data,
so
the 3D-DCT is a natural choice for object representation. Given a sequence of normalized
object image regions $\mathcal{F}=\left(f_{\mathbf{III}}(x, y, z)\right)_{N_{1}\times N_{2} \times N_{3}}$ from previous frames and
a candidate image region $\left(\tau(x, y)\right)_{N_{1}\times N_{2}}$ in the current frame, we have a new image
sequence $\mathcal{F}^{'}=\left(f_{\mathbf{III}}(x, y, z)\right)_{N_{1}\times N_{2} \times (N_{3}+1)}$ where
the first $N_{3}$ images correspond to $\mathcal{F}$ and
the last image (i.e., the $(N_{3}+1)$th image) is $\left(\tau(x, y)\right)_{N_{1}\times N_{2}}$.
According to Equ.~\eqref{eq:F_tensor_recons},
$\mathcal{F}^{'}$ can be expressed as: \vspace{-0.1cm}
\begin{equation}
\mathcal{F}^{'} = \mathbf{C}^{'}_{\mathbf{III}}\times_{1}\mathbf{A}_{1}^{T}\times_{2}\mathbf{A}^{T}_{2}\times_{3}(\mathbf{A}^{'}_{3})^{T},
\label{eq:F_new_reconstruction} \vspace{-0.1cm}
\end{equation}
where $\mathbf{C}^{'}_{\mathbf{III}}\in \mathcal{R}^{N_{1}\times N_{2} \times (N_{3}+1)}$ is the 3D-DCT coefficient matrix:
$\mathbf{C}^{'}_{\mathbf{III}} =  \mathcal{F}^{'} \times_{1}\mathbf{A}_{1}\times_{2}\mathbf{A}_{2}\times_{3}\mathbf{A}^{'}_{3}$
and $\mathbf{A}^{'}_{3} \in \mathcal{R}^{(N_{3}+1)\times (N_{3}+1)}$
is a cosine basis matrix whose entry is defined as: \vspace{-0.1cm}
\begin{equation}
a^{'}_{3}(w, z) = \left\{
\begin{array}{ll}
\sqrt{\frac{1}{N_{3}+1}}, & \mbox{if} \thickspace w=0;\\
\sqrt{\frac{2}{N_{3}+1}}\cos\left[\frac{\pi(2z+1)w}{2(N_{3}+1)}\right], & \mbox{otherwise}.
\end{array}
\right.
\label{eq:A3} \vspace{-0.1cm}
\end{equation}
According to the  properties of the 3D-DCT, the larger the values of $(u, v, w)$ are,
the higher frequency the corresponding elements of $\mathbf{C}^{'}_{\mathbf{III}}$ encode.
Usually,
the high-frequency coefficients are sparse while the low-frequency coefficients are relatively dense.
Recently, PCA (principal component analysis) tracking~\cite{Limy-Ross17} builds
a compact subspace model which maintains a set of
principal eigenvectors controlling the degree of structural information preservation.
Inspired by PCA tracking~\cite{Limy-Ross17}, we compress the 3D-DCT object representation
by retaining the relatively low-frequency
elements of $\mathbf{C}^{'}_{\mathbf{III}}$ around the origin, i.e.,
$\{(u,v,w)|u\leq \delta_{u}, v\leq \delta_{v}, w\leq \delta_{w}\}$.
As a result, we can obtain a compact 3D-DCT coefficient matrix
$C^{\ast}_{\mathbf{III}}$. Then, $\mathcal{F}^{'}$ can be approximated by: \vspace{-0.1cm}
\begin{equation}
\mathcal{F}^{'} \approx  \mathcal{F}^{\ast} = \mathbf{C}^{\ast}_{\mathbf{III}}\times_{1}\mathbf{A}_{1}^{T}\times_{2}\mathbf{A}^{T}_{2}\times_{3}(\mathbf{A}^{'}_{3})^{T}.
\label{eq:Compact_representation} \vspace{-0.1cm}
\end{equation}
Let $\mathcal{F}^{\ast}=\left(f^{\ast}_{\mathbf{III}}(x, y, z)\right)_{N_{1}\times N_{2} \times (N_{3}+1)}$ denote the corresponding reconstructed
image sequence of $\mathcal{F}^{'}$.
The loss of high frequency components introduces a reconstruction error
$\|\tau - f^{\ast}_{\mathbf{III}}(:,:,N_{3}+1)\|$, which forms the basis of the likelihood measure, as shown in Section~\ref{sec:likelihood_eval}.

\begin{algorithm}[t]
  \caption{
  Incremental 3D-DCT for object representation.
  }
  \small
  \KwIn
  {
    \begin{itemize}
          \item Cosine basis matrices $\mathbf{A}_{1}$ and $\mathbf{A}_{2}$ (whose values are fixed given $N_{1}$ and $N_{2}$)
      \item Cosine basis matrices $\mathbf{A}^{'}_{3}$
      \item New image $\left(\tau(x, y)\right)_{N_{1}\times N_{2}}$
      \item $\mathbf{D} = \mathcal{F} \times_{1}\mathbf{A}_{1}\times_{2}\mathbf{A}_{2}$ of the previous image sequence $\mathcal{F}=\left(f_{\mathbf{III}}(x, y, z)\right)_{N_{1}\times N_{2}\times N_{3}}$
    \end{itemize}
  }
  \Begin
  {
    \begin{enumerate}
\item Use the FFT to efficiently compute the 2D-DCT of $\tau$; 
\item Update $\mathbf{D}^{'}$ according to Equ.~\eqref{eq:D_update};
\item Employ the FFT to efficiently obtain the 1D-DCT of $\mathbf{D}^{'}$ along the third dimension. 
\end{enumerate}
  }

  \KwOut
  {
    \begin{itemize}
      \item 3D-DCT (i.e., $\mathbf{C}^{'}_{\mathbf{III}}$) of the current image sequence $\mathcal{F}^{'}=\left(f_{\mathbf{III}}(x, y, z)\right)_{N_{1}\times N_{2} \times (N_{3}+1)}$
    \end{itemize}
  }
  \label{alg:incremental_3D_DCT}
\end{algorithm}

\vspace{-0.1cm}
\subsection{Incremental 3D-DCT }
\vspace{-0.1cm}
\label{sec:incremental_3D_dct}

Given a sequence of training images, we have shown how to use the
3D-DCT to represent an object for visual tracking, in Equ.~\eqref{eq:Compact_representation}. As the object's appearance changes with time, it is also necessary to update the object representation.
Consequently, we propose an incremental 3D-DCT algorithm which can efficiently update the 3D-DCT based object representation as new data
arrive.

Given a new image $\left(\tau(x, y)\right)_{N_{1}\times N_{2}}$ and the transform coefficient matrix
$\mathbf{D} =  \mathcal{F} \times_{1}\mathbf{A}_{1}\times_{2}\mathbf{A}_{2}\in \mathcal{R}^{{N_{1}\times N_{2} \times N_{3}}}$ of previous
images $\mathcal{F}=\left(f_{\mathbf{III}}(x, y, z)\right)_{N_{1}\times N_{2} \times N_{3}}$,
the incremental 3D-DCT algorithm aims to efficiently compute the 3D-DCT coefficient matrix
$\mathbf{C}^{'}_{\mathbf{III}}\in \mathcal{R}^{{N_{1}\times N_{2} \times (N_{3}+1)}}$ of the previous images with the current image appended:
$\mathcal{F}^{'}=\left(f_{\mathbf{III}}(x, y, z)\right)_{N_{1}\times N_{2} \times (N_{3}+1)}$ with the last image being $\left(\tau(x, y)\right)_{N_{1}\times N_{2}}$.
Mathematically, $\mathbf{C}^{'}_{\mathbf{III}}$ is formulated as: \vspace{-0.1cm}
\begin{equation}
\mathbf{C}^{'}_{\mathbf{III}} =  \mathcal{F}^{'} \times_{1}\mathbf{A}_{1}\times_{2}\mathbf{A}_{2}\times_{3}\mathbf{A}^{'}_{3},
\label{eq:incremental_coef_mat} \vspace{-0.1cm}
\end{equation}
where $\mathbf{A}^{'}_{3}\in \mathcal{R}^{(N_{3}+1)\times (N_{3}+1)}$
is referred to in Equ.~\eqref{eq:F_new_reconstruction}.
In principle,  Equ.~\eqref{eq:incremental_coef_mat} can be computed in the following two stages:
1) compute the 2D-DCT coefficients for each image, i.e., $\mathbf{D}^{'} = \mathcal{F}^{'} \times_{1}\mathbf{A}_{1}\times_{2}\mathbf{A}_{2}$;
and 2) calculate the 1D-DCT coefficients along the time dimension, i.e, $\mathbf{C}^{'}_{\mathbf{III}} = \mathbf{D}^{'}\times_{3}\mathbf{A}^{'}_{3}$.

\begin{figure}[t]
\vspace{-0.05cm}
\begin{center}
   \includegraphics[width=1\linewidth]{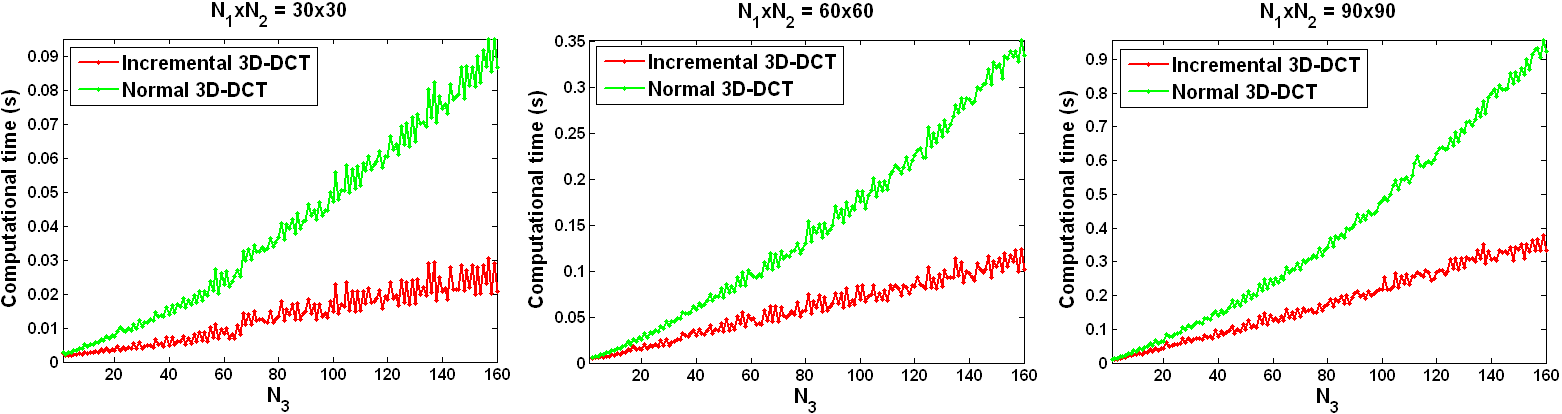}
\end{center}
\vspace{-0.6cm}
   \caption{Comparison on the computational time between the normal 3D-DCT and our incremental 3D-DCT. The three subfigures correspond to
   different configurations of $N_{1}\times N_{2}$ (i.e., $30\times30$, $60\times60$, and $90\times90$). In each subfigure, the x-axis is associated with $N_{3}$; the y-axis corresponds to the computational time.
   Clearly, as $N_{3}$ increases, the computational time of the normal 3D-DCT grows much faster than that of the incremental 3D-DCT.
   }
    \label{fig:Incremental3D-DCT} \vspace{-0.6cm}
\end{figure}

According to the definition of the 3D-DCT, the CBF matrices $\mathbf{A}_{1}$ and $\mathbf{A}_{2}$ only depend on the row and column dimensions (i.e., $N_{1}$ and $N_{2}$),
respectively. Since both $N_{1}$ and $N_{2}$ are unchanged during visual tracking, both $\mathbf{A}_{1}$ and $\mathbf{A}_{2}$ remain constant.
In addition, $\mathcal{F}^{'}$ is a concatenation of $\mathcal{F}$ and $\left(\tau(x, y)\right)_{N_{1}\times N_{2}}$ along the third dimension.
According to the property of tensor algebra, $\mathbf{D}^{'}$ can be decomposed as:
\begin{equation}
\mathbf{D}^{'}(:,:,k) = \left\{
\begin{array}{ll}
\mathbf{D}(:,:,k), & \mbox{if} \thickspace 1\leq k \leq N_{3};\\
\tau\times_{1}\mathbf{A}_{1}\times_{2}\mathbf{A}_{2}, & k=N_{3}+1;
\end{array}
\right.
\label{eq:D_update}
\end{equation}
Given $\mathbf{D}$,  $\mathbf{D}'$ can be efficiently updated by only computing the term $\tau\times_{1}\mathbf{A}_{1}\times_{2}\mathbf{A}_{2}$.
Moreover, $\mathbf{A}^{'}_{3}$ is only dependent on the variable $N_{3}$. Once $N_{3}$ is fixed, $\mathbf{A}^{'}_{3}$ is also fixed.
In addition, $\tau\times_{1}\mathbf{A}_{1}\times_{2}\mathbf{A}_{2}$ can be viewed as the 2D-DCT along the first two dimensions (i.e., $x$ and $y$);
and
$\mathbf{C}^{'}_{\mathbf{III}} = \mathbf{D}^{'}\times_{3}\mathbf{A}^{'}_{3}$ can be viewed as the 1D-DCT along the time dimension.
To further reduce the computational time of the 1D-DCT and 2D-DCT, we employ a fast algorithm using the Fast Fourier Transform (FFT)
to efficiently compute the DCT and its inverse~\cite{AKJ-DCT,khayam-tr2003}. The complete procedure of the incremental 3D-DCT algorithm is summarized in
Algorithm~\ref{alg:incremental_3D_DCT}.

The complexity of our incremental
algorithm is $O(N_1 N_2(\log N_1 + \log N_2) + N_1 N_2 N_3 \log N_3)$ at each frame.
In contrast, using a traditional batch-mode strategy for DCT computation, the complexity of the normal 3D-DCT algorithm becomes
$O(N_1 N_2 N_3 (\log N_1 + \log N_2 + \log N_3))$.
To illustrate the computational efficiency of the incremental 3D-DCT algorithm, Fig.~\ref{fig:Incremental3D-DCT}
shows the computational time of the incremental 3D-DCT and normal 3D-DCT algorithms for different values of $N_{1}$, $N_{2}$, and $N_{3}$.
Although the computation time of both algorithms increases with $N_3$, the growth rate of the incremental 3D-DCT algorithm is much lower.

\begin{algorithm}[t!]
\small
\caption{Incremental 3D-DCT  object tracking.}
\label{alg:Framwork}
  \KwIn
  { New frame $t$,
 previous object state $\mathbf{Z}_{t-1}^{\ast}$,
 previous positive and negative sample sets: $\mathcal{F}_{+} = \left(f^{+}_{\mathbf{III}}(x, y, z)\right)_{N_{1}\times N_{2} \times N^{+}_{3}}$
 and $\mathcal{F}_{-} = \left(f^{-}_{\mathbf{III}}(x, y, z)\right)_{N_{1}\times N_{2} \times N^{-}_{3}}$,
maximum buffer size $\mathbb{T}$.
 }
\textbf{Initialization:}\\
-- $t=1$.\\
-- Manually set the initial object state $\mathbf{Z}_{t}^{\ast}$.\\
-- Collect positive (or negative) samples to form training sets
$\mathcal{F}_{+}=\mathbb{Z}^{+}_{t}$ and
$\mathcal{F}_{-}=\mathbb{Z}^{-}_{t}$ (see
    Section \ref{sec:training_sample_selection}).\\
  \Begin
  {
  \begin{itemize}
\item Sample $V$ candidate object states
    $\{\mathbf{Z}_{tj}\}_{j=1}^{V}$ according to Equ.~\eqref{eq:particle_filter}.
\item Crop out the corresponding image regions $\{o_{tj}\}_{j=1}^{V}$ of $\{\mathbf{Z}_{tj}\}_{j=1}^{V}$.
\item Resize each candidate image region $o_{tj}$ to $N_{1}\times N_{2}$ pixels.
\item \For{\emph{each} $\mathbf{Z}_{tj}$}
    {
     \begin{enumerate}
         \item Find the $K$ nearest neighbors $\mathcal{F}_{+}^{K}\in \mathcal{R}^{N_{1}\times N_{2} \times K}$
          (or $\mathcal{F}_{-}^{K}\in \mathcal{R}^{N_{1}\times N_{2} \times K}$) of
          a candidate\\ sample $\tau$ (i.e., $\tau = o_{tj}$)  from $\mathcal{F}_{+}$ (or $\mathcal{F}_{-}$).
          \item Obtain the 3D signals $\mathcal{F}_{+}^{'}$ and
              $\mathcal{F}_{-}^{'}$ through the concatenations of
              $(\mathcal{F}_{+}^{K}, \tau)$ and $(\mathcal{F}_{-}^{K},
              \tau)$.
          \item Perform the incremental 3D-DCT in Algorithm~\ref{alg:incremental_3D_DCT} to compute the 3D-DCT
              coefficient matrices:  $\mathbf{C}^{'}_{\mathbf{III}_{+}}$
                and $\mathbf{C}^{'}_{\mathbf{III}_{-}}$.
          \item Compute the compact 3D-DCT coefficient matrices $\mathbf{C}^{\ast}_{\mathbf{III}_{+}}$
                and $\mathbf{C}^{\ast}_{\mathbf{III}_{-}}$ by discarding the 
                high-frequency coefficients of  
                \\
                $\mathbf{C}^{'}_{\mathbf{III}_{+}}$
                and $\mathbf{C}^{'}_{\mathbf{III}_{-}}$.
          \item Calculate the reconstructed representations of  $\mathcal{F}_{+}^{'}$ and $\mathcal{F}_{-}^{'}$
                as $\mathcal{F}^{\ast}_{+}$ and
                $\mathcal{F}^{\ast}_{-}$ by Equ.~\eqref{eq:Compact_representation}.
          \item Compute the reconstruction likelihoods
              $\mathcal{L}_{\tau_{+}}$ and $\mathcal{L}_{\tau_{-}}$
              using Equ.~\eqref{eq:Middle_likelihood}.
          \item Calculate the final likelihood
              $\mathcal{L}^{\ast}_{\tau}$ using Equ.~\eqref{eq:final_likelihood}.
     \end{enumerate}
     }
\item Determine the optimal object state $\mathbf{Z}_{t}^{\ast}$ by
    the MAP estimation (referred to in Equ.~\eqref{eq:map_estimation}).
\item Select positive (or negative) samples $\mathbb{Z}^{+}_{t}$ (or $\mathbb{Z}^{-}_{t}$) (referred to in Sec.~\ref{sec:training_sample_selection}).
\item Update the training sample sets $\mathcal{F}_{+}$ and $\mathcal{F}_{-}$ with $\mathcal{F}_{+} \bigcup\mathbb{Z}^{+}_{t}$
and $\mathcal{F}_{-}\bigcup\mathbb{Z}^{-}_{t}$.
\item $N^{+}_{3}=N^{+}_{3}+|\mathbb{Z}^{+}_{t}|$ and $N^{-}_{3}=N^{-}_{3}+|\mathbb{Z}^{-}_{t}|$.
\item Maintain the positive and negative sample sets as follows:
\begin{itemize}
\item If $N^{+}_{3}> \mathbb{T}$, then $\mathcal{F}_{+}$ is truncated to keep the last $\mathbb{T}$ elements.
\item If $N^{-}_{3}> \mathbb{T}$, then $\mathcal{F}_{-}$ is truncated to keep the last $\mathbb{T}$ elements.
\end{itemize}
\end{itemize}
}
  \KwOut
  { Current object state $\mathbf{Z}_{t}^{\ast}$, updated positive and negative sample sets $\mathcal{F}_{+}$ and $\mathcal{F}_{-}$.
}
\end{algorithm}

\section{Incremental 3D-DCT based tracking  \label{sec:incremental_tracking_alg}}

In this section, we propose a complete 3D-DCT based tracking algorithm, which is composed of three main modules:
\begin{itemize}
\item {\em training sample selection}: select positive and negative samples for discriminative learning;
\item {\em likelihood evaluation}: compute the similarity between candidate samples and the
3D-DCT based observation model;
\item {\em motion estimation}:  generate  candidate samples and estimate the  object state.
\end{itemize}
Algorithm~\ref{alg:Framwork} lists the workflow of the proposed
tracking algorithm. Next, we will discuss the three modules in detail.

\subsection{Training sample selection \label{sec:training_sample_selection}}

\begin{figure}[t]
\vspace{-0.05cm}
\begin{center}
   \includegraphics[width=1\linewidth]{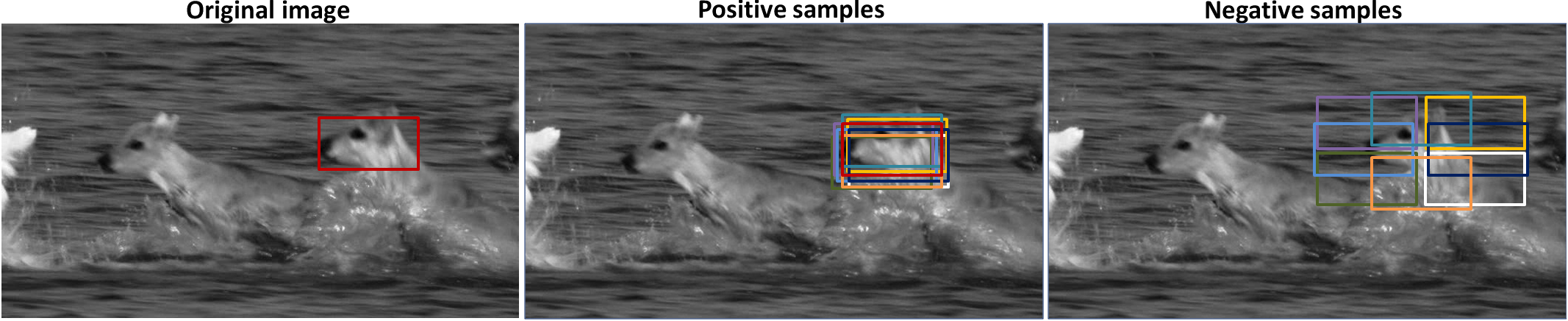}
\end{center}
\vspace{-0.66cm}
   \caption{Illustration of training sample selection. The left subfigure plots the bounding box corresponding to the current tracker location;
   the middle subfigure shows the selected positive samples; and the right subfigure displays the selected negative samples. Different colors are assoicated
   with different samples.}
    \label{fig:training_sample_selection} \vspace{-0.8cm}
\end{figure}

Similar to~\cite{Babenko-Yang-Belongie-cvpr2009}, we take a spatial distance-based strategy for training sample selection.
Namely, the image regions from a small neighborhood around the object location
are selected as positive samples, while the negative samples are generated by selecting the
image regions which are relatively far from the object location.
Specifically, we draw a number of samples $\mathbb{Z}_{t}$ from Equ.~\eqref{eq:particle_filter}, and then
an ascending sort for the samples from $\mathbb{Z}_{t}$ is made according to their
spatial distances to the current object location, resulting in a sorted sample set $\mathbb{Z}_{t}^{s}$.
By selecting the first few samples from $\mathbb{Z}_{t}^{s}$, we have
a subset $\mathbb{Z}^{+}_{t}$ that is the final positive sample set, as shown in the middle part of Fig.~\ref{fig:training_sample_selection}.
The negative sample set $\mathbb{Z}^{-}_{t}$ is generated in the area around the current tracker location, as shown in the right part of Fig.~\ref{fig:training_sample_selection}.

\subsection{Likelihood evaluation}
\label{sec:likelihood_eval}

During tracking, each of positive and negative samples is
normalized to $N_{1}\times N_{2}$ pixels. Without loss of generality, we assume the numbers of the positive and
negative samples to be  $N^{+}_{3}$ and $N^{-}_{3}$. The positive and negative sample sequences are denoted as
$\mathcal{F}_{+} = \left(f^{+}_{\mathbf{III}}(x, y, z)\right)_{N_{1}\times N_{2} \times N^{+}_{3}}$
and $\mathcal{F}_{-} = \left(f^{-}_{\mathbf{III}}(x, y, z)\right)_{N_{1}\times N_{2} \times N^{-}_{3}}$, respectively.
Based on $\mathcal{F}_{+}$ and
$\mathcal{F}_{-}$, we
evaluate the likelihood of
a candidate sample
$\left(\tau(x, y)\right)_{N_{1}\times N_{2}}$
belonging to the foreground object.
Since  the
appearance of  $\mathcal{F}_{+}$ and $\mathcal{F}_{-}$ is likely to  vary significantly
as time progresses, it is not necessary for the 3D-DCT to use all samples in  $\mathcal{F}_{+}$ and $\mathcal{F}_{-}$
to represent the candidate sample $\left(\tau(x, y)\right)_{N_{1}\times N_{2}}$.
As pointed out by~\cite{Wang-Yang-Yu-Lv-Huang-Gong-CVPR2010},
 locality is more essential than sparsity because
locality usually results in sparsity but not necessarily vice versa.
As a result, a locality-constrained strategy is taken to construct a compact object representation using the proposed incremental 3D-DCT algorithm.

\begin{figure}[t]
\vspace{-0.13cm}
\begin{center}
 \includegraphics[width=0.78\linewidth]{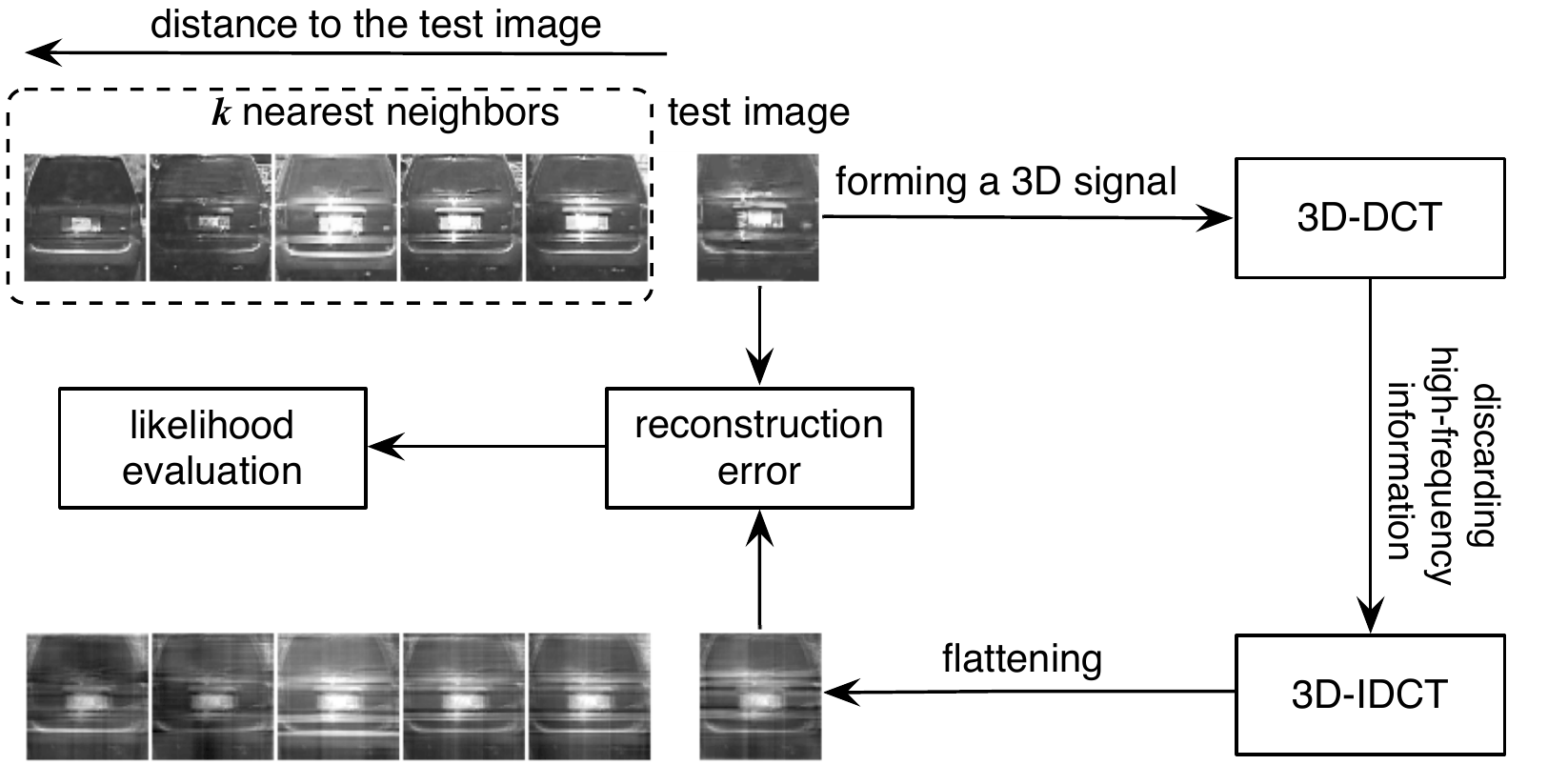}\\
\end{center}
\vspace{-0.8cm}
   \caption{Illustration of the process of computing the reconstruction likelihood between test images and
     training images using the 3D-DCT and 3D-IDCT.}
    \label{fig:manu_Reconstruction_example} \vspace{-0.35cm}
\end{figure}

\begin{figure}[h!]
\vspace{-0.05cm}
\begin{center}
   \includegraphics[width=0.73\linewidth]{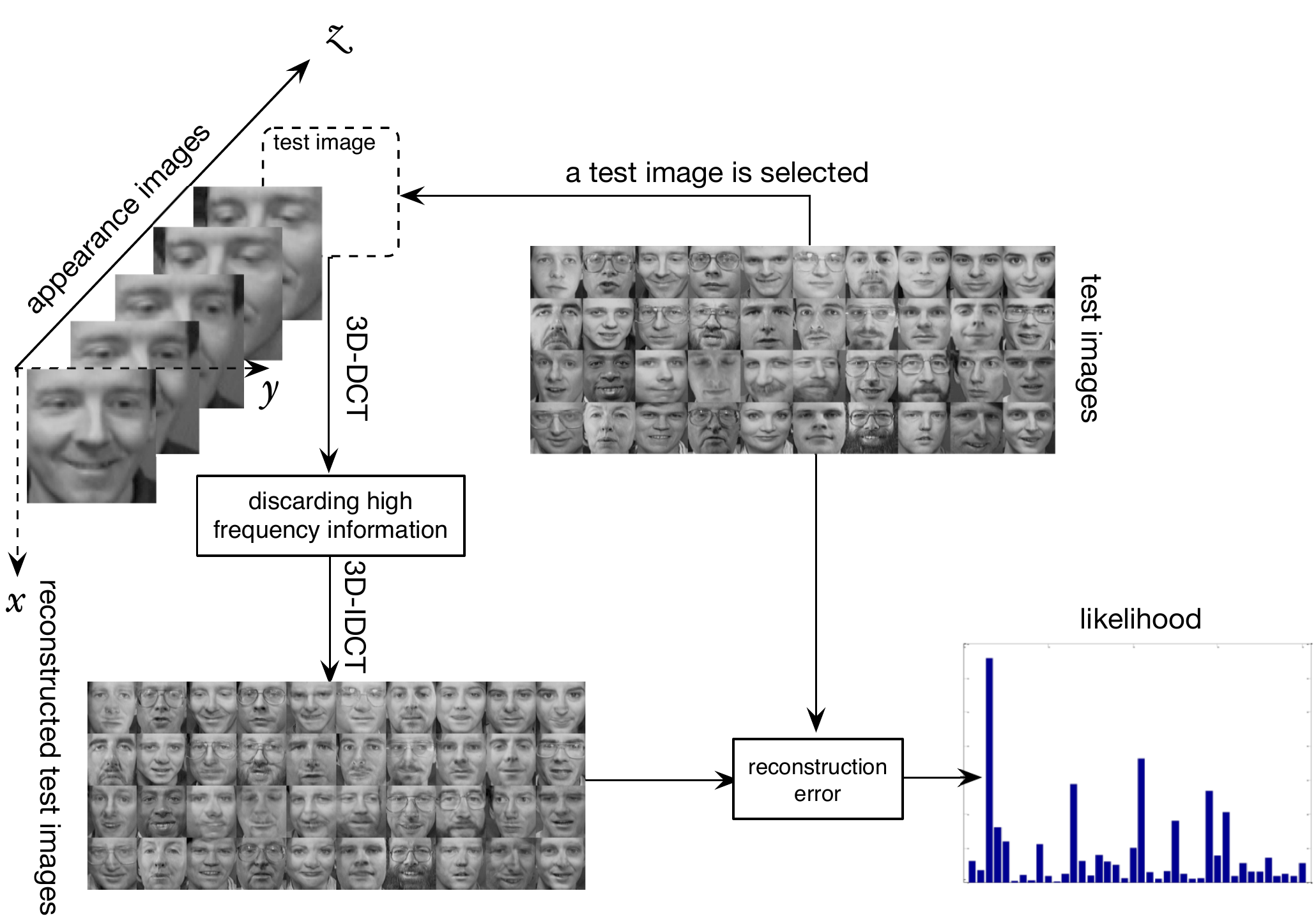}
\end{center}
\vspace{-0.68cm}
   \caption{Example of computing the likelihood scores between test images and training images. The left part
   shows the training image sequence; the top-right part displays the test images; the bottom-middle part exhibits
   the reconstructed images by 3D-DCT and 3D-IDCT; the bottom-right part plots the corresponding likelihood scores (computed
   by Equ.~\eqref{eq:Middle_likelihood}).}
    \label{fig:manu_Reconstruction_example2} \vspace{-0.6cm}
\end{figure}

Specifically, we first compute the $K$-nearest neighbors (referred to as $\mathcal{F}_{+}^{K}\in \mathcal{R}^{N_{1}\times N_{2} \times K}$ and $\mathcal{F}_{-}^{K}\in \mathcal{R}^{N_{1}\times N_{2} \times K}$)
of the candidate sample $\tau$ from $\mathcal{F}_{+}$ and $\mathcal{F}_{-}$, sort them by their sum-squared distance to $\tau$ (as shown in the top-left part
of Fig.~\ref{fig:manu_Reconstruction_example}),
and then utilize the incremental 3D-DCT algorithm to construct the compact object representation.
Let $\mathcal{F}_{+}^{'}$ and $\mathcal{F}_{-}^{'}$ denote the concatenations of $(\mathcal{F}_{+}^{K}, \tau)$ and $(\mathcal{F}_{-}^{K}, \tau)$, respectively.
Through the incremental 3D-DCT algorithm, the corresponding 3D-DCT coefficient matrices $\mathbf{C}^{'}_{\mathbf{III}_{+}}$
and $\mathbf{C}^{'}_{\mathbf{III}_{-}}$ can be efficiently calculated. After discarding the high-frequency coefficients, we can
obtain the corresponding compact 3D-DCT coefficient matrices $\mathbf{C}^{\ast}_{\mathbf{III}_{+}}$
and $\mathbf{C}^{\ast}_{\mathbf{III}_{-}}$. Based on Equ.~\eqref{eq:Compact_representation}, the reconstructed representations of  $\mathcal{F}_{+}^{'}$ and $\mathcal{F}_{-}^{'}$
are obtained as $\mathcal{F}^{\ast}_{+}$ and $\mathcal{F}^{\ast}_{-}$, respectively.
We compute the following reconstruction likelihoods:
\vspace{-0.1cm}
\begin{equation}
\begin{array}{l}
\mathcal{L}_{\tau_{+}} = \exp\left(-\frac{1}{2\gamma^{2}_{+}}\|\tau - f^{\ast}_{\mathbf{III}_{+}}(:,:,K+1)\|^{2}\right), \vspace{0.15cm}\\
\mathcal{L}_{\tau_{-}} = \exp\left(-\frac{1}{2\gamma^{2}_{-}}\|\tau - f^{\ast}_{\mathbf{III}_{-}}(:,:,K+1)\|^{2}\right),
\end{array}
\label{eq:Middle_likelihood} \vspace{-0.1cm}
\end{equation}
where $\gamma_{+}$ and $\gamma_{-}$ are two scaling factors, $f^{\ast}_{\mathbf{III}_{+}}(:,:,K+1)$
and $f^{\ast}_{\mathbf{III}_{-}}(:,:,K+1)$ are respectively the last images of $\mathcal{F}^{\ast}_{+}$ and $\mathcal{F}^{\ast}_{-}$.
Figs.~\ref{fig:manu_Reconstruction_example} and \ref{fig:manu_Reconstruction_example2}
illustrates the process of computing the reconstruction likelihood between test samples and training samples (i.e., car and face samples) using the 3D-DCT and 3D-IDCT.
Based on $\mathcal{L}_{\tau_{+}}$ and $\mathcal{L}_{\tau_{-}}$,
we define the final likelihood evaluation criterion: \vspace{-0.16cm}
\begin{equation}
\begin{array}{ll}
\mathcal{L}^{\ast}_{\tau} &= \rho\left(\mathcal{L}_{\tau_{+}}-\lambda\mathcal{L}_{\tau_{-}}\right)\\
\end{array}
\label{eq:final_likelihood} \vspace{-0.16cm}
\end{equation}
where $\lambda$ is a weight factor and $\rho(x)=\frac{1}{1+\exp(-x)}$ is the sigmoid function.

\begin{figure*}[h!]
\begin{center}
   \includegraphics[width=0.6\linewidth]{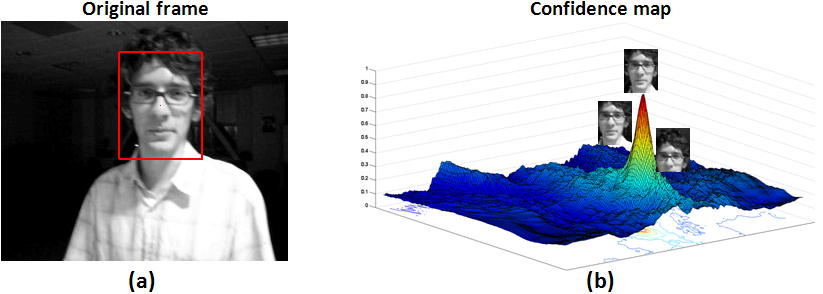}
\end{center}
\vspace{-0.85cm}
   \caption{{Demonstration of the discriminative ability of the 3D-DCT based object representation used by our tracker.
   (a) shows the original frame; and (b) displays a confidence map, each element of which corresponds to an image
   patch in the entire image search space.
    }}
    \label{fig:confidence_map} \vspace{-0.7cm}
\end{figure*}

To demonstrate the discriminative ability of the proposed 3D-DCT based observation model, we plot a confidence map defined in the entire image search space
(shown in Fig.~\ref{fig:confidence_map}(a)).
Each element of the confidence map is computed by
measuring the likelihood score of the candidate bounding box centered
at this pixel belonging to the learned observation model, according to
Equ.~\eqref{eq:final_likelihood}.
For better visualization,  $\mathcal{L}^{\ast}_{\tau}$ is normalized to [0, 1].
After calculating all the normalized likelihood scores at different locations, we have a confidence map which is
shown in Fig.~\ref{fig:confidence_map}(b).
From Fig.~\ref{fig:confidence_map}(b), we can see that
the confidence map has an obvious uni-modal peak, which indicates that
the proposed observation model has a good discriminative ability in this image.

\subsection{Motion estimation \label{sec:Motion_model}}

The motion estimation module is based on a particle filter~\cite{Isard-Blake-ECCV1996} that is
a Markov model with hidden state variables. The particle filter
can be divided into the prediction and the update steps: \vspace{-0.2cm}
\[
 p(\mathbf{Z}_{t}\hspace{-0.05cm}\mid \hspace{-0.05cm}\mathcal
{O}_{t-1} )\hspace{-0.0cm}\propto \int
\hspace{-0.1cm}p(\mathbf{Z}_{t}\hspace{-0.1cm}\mid
\hspace{-0.1cm}\mathbf{Z}_{t-1})p(\mathbf{Z}_{t-1}\hspace{-0.1cm}\mid \hspace{-0.1cm}
\mathcal {O}_{t-1} )d\mathbf{Z}_{t-1}, \hspace{-0.25cm}
\label{eq:prediction_filter} \vspace{-0.26cm}
\]
\[
p(\mathbf{Z}_{t}\hspace{-0.1cm}\mid \hspace{-0.1cm}\mathcal
{O}_{t} )\hspace{-0.0cm}\propto\hspace{-0.0cm}
p(o_{t}\hspace{-0.1cm}\mid \hspace{-0.1cm}\mathbf{Z}_{t})p(\mathbf{Z}_{t}\hspace{-0.1cm}\mid \hspace{-0.1cm}\mathcal
{O}_{t-1} ),\hspace{-0.0cm}
\label{eq:update_filter} \vspace{-0.26cm}
\]
where $\mathcal
{O}_{t}=\{o_{1},\ldots,o_{t}\}$ are observation variables, $p(o_{t}\hspace{-0.1cm}\mid \hspace{-0.1cm}\mathbf{Z}_{t})$ denotes the
observation model, and $p(\mathbf{Z}_{t}\hspace{-0.1cm}\mid
\hspace{-0.1cm}\mathbf{Z}_{t-1})$ represents the state transition model.
For the sake of computational efficiency, we only consider
the motion information in translation and scaling.
Specifically, let $\mathbf{Z}_{t}= (\mathcal{X}_{t}, \mathcal{Y}_{t}, \mathcal{S}_{t})$
denote the motion parameters including
$\mathcal{X}$ translation, $\mathcal{Y}$ translation, and scaling.
The motion model between two consecutive frames is assumed to be a Gaussian distribution:
\vspace{-0.33cm}
\begin{equation}
p(\mathbf{Z}_{t}|\mathbf{Z}_{t-1}) = \mathcal{N}(\mathbf{Z}_{t}; \mathbf{Z}_{t-1}, \Sigma),
\label{eq:particle_filter}
\vspace{-0.33cm}
\end{equation}
where $\Sigma$ denotes a diagonal covariance matrix with diagonal
elements: $\sigma_{\mathcal{X}}^{2}$, $\sigma_{\mathcal{Y}}^{2}$, and $\sigma_{\mathcal{S}}^{2}$.
For each state $\mathbf{Z}_{t}$, there is a corresponding image region $o_{t}$ that is normalized to $N_{1}\times N_{2}$ pixels by image scaling.
 The likelihood $p(o_{t}\hspace{-0.1cm}\mid \hspace{-0.1cm}\mathbf{Z}_{t})$ is defined as:
$p(o_{t}\hspace{-0.1cm}\mid \hspace{-0.1cm}\mathbf{Z}_{t}) \propto
\mathcal{L}^{\ast}_{\tau}$
where $\mathcal{L}^{\ast}_{\tau}$ is defined in
Equ.~\eqref{eq:final_likelihood}.
Thus, the optimal object state $\mathbf{Z}_{t}^{\ast}$ at time $t$ can be determined by solving the following maximum a posterior (MAP) problem: \vspace{-0.28cm}
\begin{equation}
\mathbf{Z}_{t}^{\ast} = \underset{\mathbf{Z}_{t}}{\arg \max} \thinspace p(\mathbf{Z}_{t}\hspace{-0.1cm}\mid \hspace{-0.1cm}\mathcal
{O}_{t}).
\label{eq:map_estimation} \vspace{-0.05cm}
\end{equation}

\section{Experiments\label{sec:experimental_result}}

\subsection{Data description and implementation details}

We evaluate the performance of the proposed
tracker (referred to as ITDT) on twenty video sequences, which are captured in different
scenes and composed of 8-bit
grayscale images.
In these video sequences, several complicated factors lead to drastic appearance changes of
the tracked objects, including illumination variation, occlusion,
out-of-plane rotation, background distraction,
small target, motion blurring, pose variation, etc. In order to verify the effectiveness of the proposed
tracker on these video sequences,
a large number of experiments are conducted. These experiments have two main goals:
to verify the robustness of the proposed ITDT in various challenging
situations, and to evaluate the adaptive capability of ITDT in tolerating complicated appearance changes.

The proposed ITDT is implemented in Matlab on a workstation
with an Intel Core 2 Duo 2.66GHz processor and 3.24G RAM.
The average running time of the proposed ITDT is about 0.8 second per frame.
During tracking, the pixels values of each frame are normalized into $[0, 1]$.
For the sake of computational efficiency, we only consider the object state information in 2D translation and
scaling in the particle filtering module, where the particle number is set to 200.
Each particle is associated with an image patch.
After image scaling, the image patch
is normalized to $N_{1}\times N_{2}$ pixels.
In the experiments, the parameters $(N_{1}, N_{2})$ are chosen as $(30, 30)$.
The scaling factors ($\gamma_{+}$, $\gamma_{-}$) in Equ.~\eqref{eq:Middle_likelihood} are both set to 1.2.
The weight factor $\lambda$ in Equ.~\eqref{eq:final_likelihood} is set to 0.1.
The number of nearest neighbors $K$ in Algorithm~\ref{alg:Framwork} is chosen as 15.
The parameter $\mathbb{T}$ (i.e., maximum buffer size)
in Algorithm~\ref{alg:Framwork}
is set to 500.
These parameter settings remain the same throughout all the experiments in the paper.
As for the user-defined tasks on different video sequences,
these parameter settings can be slightly readjusted
to achieve a better tracking performance.

\vspace{-0.25cm}
\subsection{Competing trackers}

We compare the proposed tracker with several other state-of-the-art trackers qualitatively and quantitatively.
The competing trackers are referred to as
FragT\footnote[1]{http://www.cs.technion.ac.il/$\sim$amita/fragtrack/fragtrack.htm} (Fragment-based tracker~\cite{Adam-Fragment-2006}),
MILT\footnote[2]{http://vision.ucsd.edu/$\sim$bbabenko/project$\_$miltrack.shtml} (multiple instance boosting-based
tracker~\cite{Babenko-Yang-Belongie-cvpr2009}), VTD\footnote[3]{http://cv.snu.ac.kr/research/$\sim$vtd/} (visual tracking
decomposition~\cite{Kwon-Lee-CVPR2010}), OAB\footnote[4]{http://www.vision.ee.ethz.ch/boostingTrackers/download.htm} (online
AdaBoost~\cite{Grabner-Grabner-Bischof-BMVC2006}), IPCA\footnote[5]{http://www.cs.utoronto.ca/$\sim$dross/ivt/} (incremental
PCA~\cite{Limy-Ross17}), and L1T\footnote[6]{{http://www.ist.temple.edu/$\sim$hbling}} ($\ell_{1}$ tracker~\cite{Meo-Ling-ICCV09}).
Furthermore, IPCA, VTD, and L1T make use of particle filters for state inference
while FragT, MILT, and OAB utilize the strategy of sliding window search for state inference.
We directly use the public source
codes of FragT, MILT, VTD, OAB,
IPCA,  and L1T.
In the experiments, OAB has two different versions, i.e., OAB1 and OAB5,
which utilize two different positive sample search radiuses (i.e., $r=1$ and $r=5$ selected in the same way as~\cite{Babenko-Yang-Belongie-cvpr2009})
for learning AdaBoost classifiers.

We select these seven competing trackers for the following reasons.  First,
as a recently proposed discriminant learning-based tracker,
MILT takes advantage of multiple instance boosting for object/non-object classification.
Based on the multi-instance object representation, MILT is capable of
capturing
the inherent ambiguity of object localization.
In contrast, OAB is based on online
single-instance boosting for object/non-object classification.  The goal of comparing ITDT with
MILT and OAB is to demonstrate the discriminative capabilities of ITDT in
handling large appearance variations.  In addition, based on a fragment-based object representation, FragT is capable of
fully capturing the spatial layout information of the object region, resulting
in the tracking robustness.
Based on incremental principal component analysis,
IPCA constructs an eigenspace-based observation model for visual tracking.
L1T converts the problem of visual tracking to that of sparse approximation based on
$\ell_{1}$-regularized minimization.
As a recently proposed
tracker, VTD uses sparse principal component
analysis to decompose the observation (or motion) model into a set of basic
observation (or motion) models, each of which covers a specific type of object
appearance (or motion).
Thus, comparing ITDT with FragT, IPCA,  L1T, and VTD can show their capabilities of
tolerating complicated appearance changes.

\begin{figure}[t]
\vspace{-0.05cm}
\hspace{-0.9cm}
\begin{center}
   \includegraphics[width=0.91\linewidth]{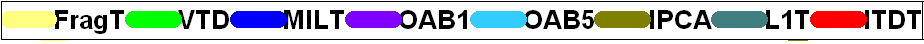}
   \includegraphics[width=0.91\linewidth]{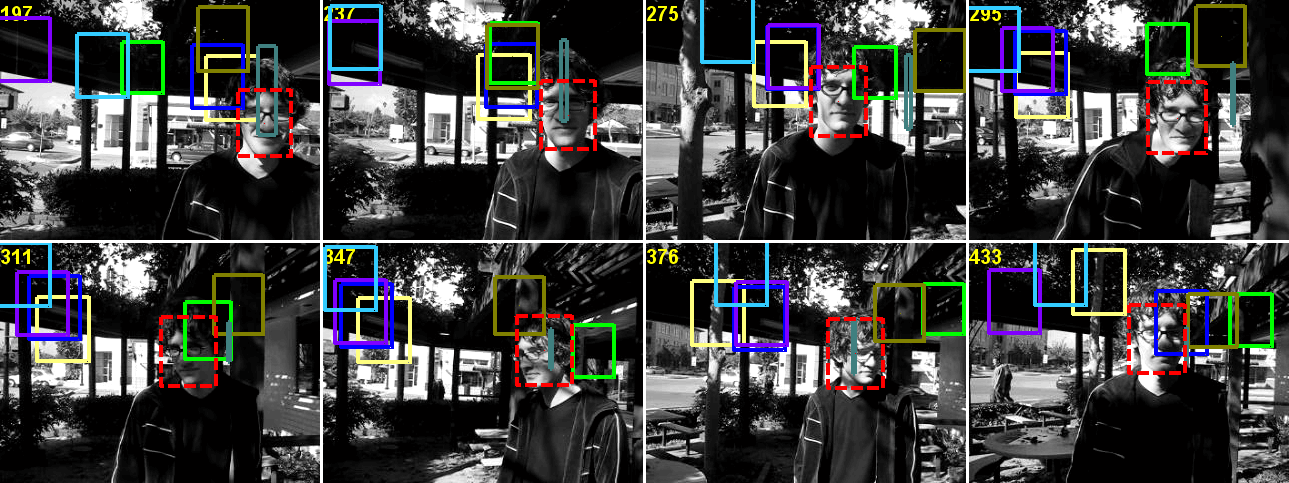}
\end{center}
\vspace{-0.68cm}
   \caption{The tracking results of the eight trackers over the
   representative frames (i.e., the 197th, 237th, 275th, 295th, 311th, 347th, 376th, and 433rd frames) of the ``\textit{trellis70}'' video sequence
   in the
   scenarios with drastic illumination changes and  head pose variations.}
    \label{fig:exp_trellis70}
    \vspace{-0.42cm}
\end{figure}

\begin{figure}[t]
\vspace{-0.05cm}
\begin{center}
\includegraphics[width=0.88\linewidth]{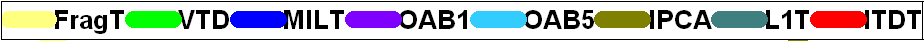}
   \includegraphics[width=0.88\linewidth]{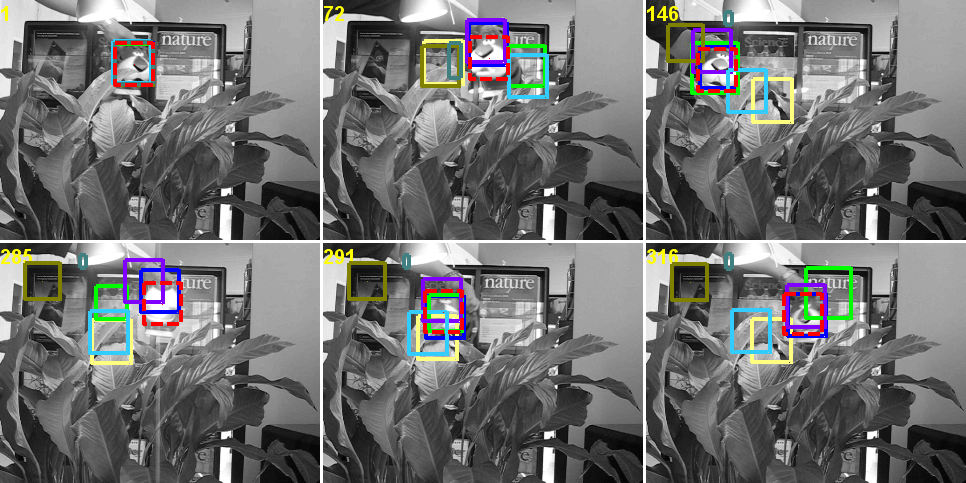}
\end{center}
\vspace{-0.8cm}
   \caption{The tracking results of the eight trackers over the
   representative frames (i.e., the 1st, 72nd, 146th, 285th, 291st, and 316th frames) of the ``\textit{tiger}'' video sequence
   in the
   scenarios with partial occlusion, illumination change, pose variation, and motion blurring.}
    \label{fig:exp_tiger}     \vspace{-0.58cm}
\end{figure}

\begin{figure}[t]
\vspace{-0.05cm}
\begin{center}
\includegraphics[width=0.88\linewidth]{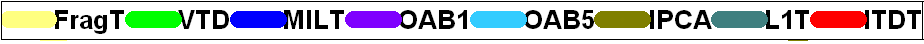}
   \includegraphics[width=0.88\linewidth]{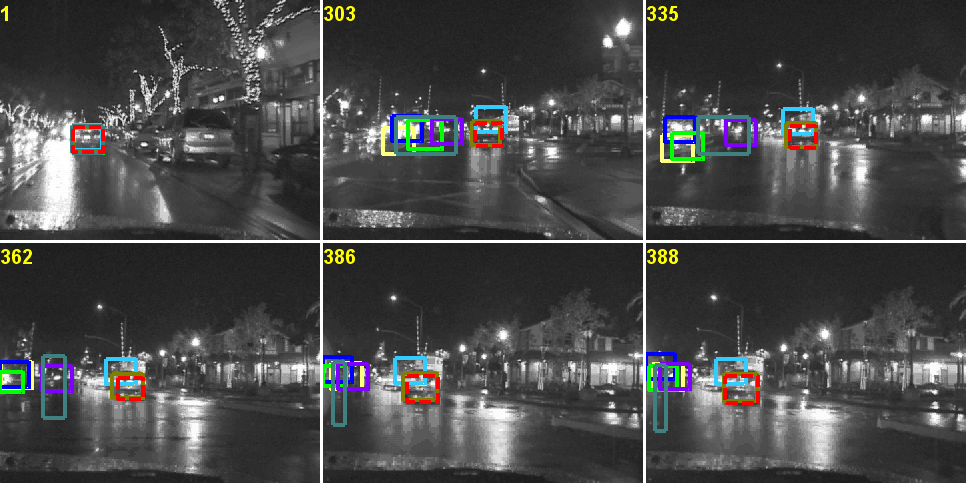}
\end{center}
\vspace{-0.8cm}
   \caption{The tracking results of the eight trackers over the
   representative frames (i.e., the 1st, 303rd, 335th, 362nd, 386th, and 388th frames) of the ``\textit{car11}'' video sequence
    in the
   scenarios with varying lighting conditions and background clutters.}
    \label{fig:exp_car11}     \vspace{-0.46cm}
\end{figure}

\begin{figure}[t]
\vspace{-0.05cm}
\begin{center}
\includegraphics[width=0.88\linewidth]{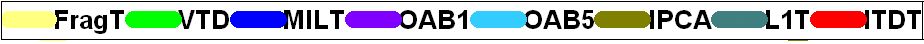}
   \includegraphics[width=0.88\linewidth]{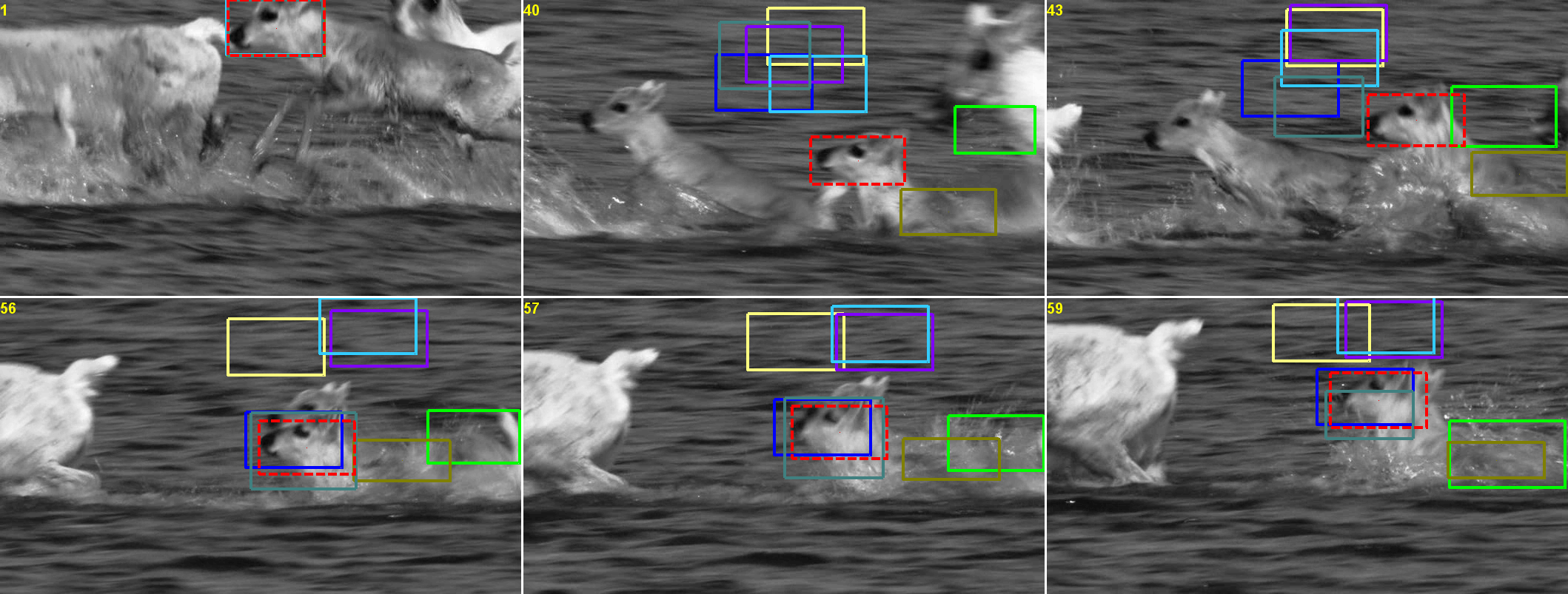}
\end{center}
\vspace{-0.8cm}
   \caption{The tracking results of the eight trackers over the
   representative frames (i.e., the 1st, 40th, 43rd, 56th, 57th, and 59th frames) of the ``\textit{animal}'' video sequence
   in the
   scenarios with motion blurring and background distraction.}
    \label{fig:exp_animal}     \vspace{-0.58cm}
\end{figure}

\begin{figure}[t]
\vspace{-0.05cm}
\begin{center}
\includegraphics[width=0.88\linewidth]{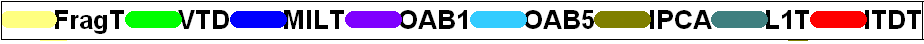}
   \includegraphics[width=0.88\linewidth]{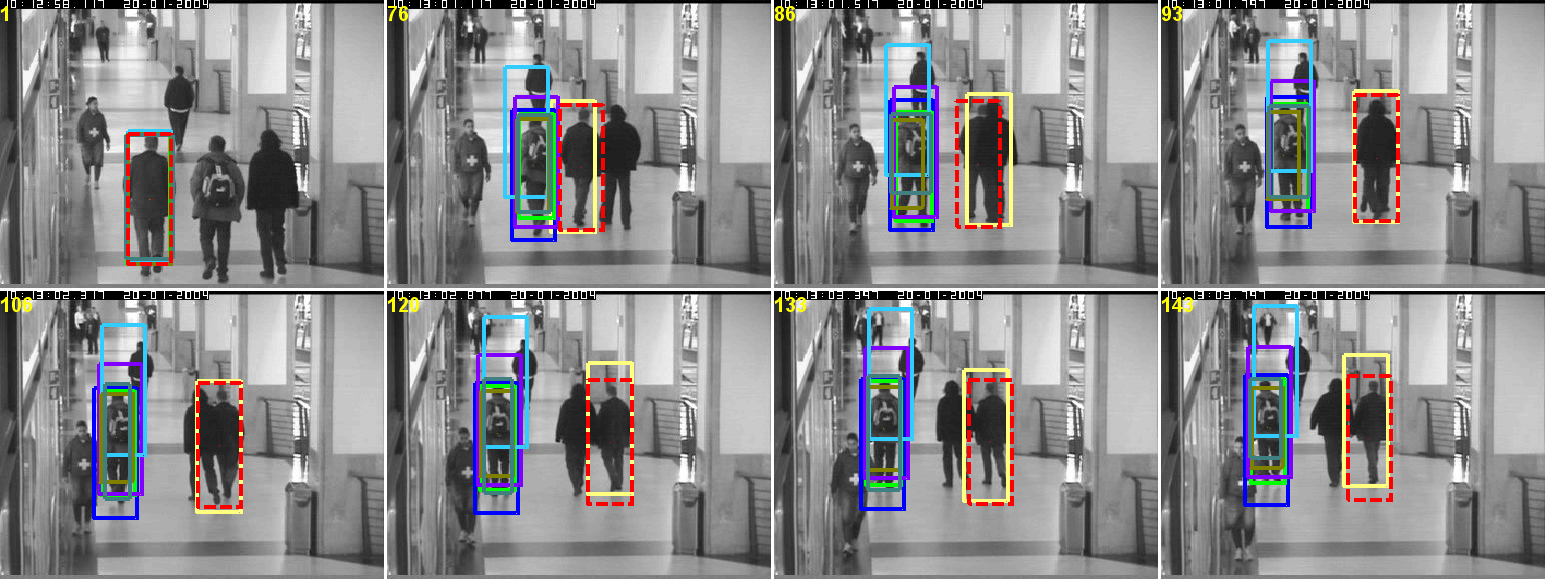}
\end{center}
\vspace{-0.8cm}
   \caption{The tracking results of the eight trackers over the
   representative frames (i.e., the 1st, 76th, 86th, 93rd, 106th, 120th, 133rd, and 143rd frames)
   of the ``\textit{sub-three-persons}'' video sequence
   in the
   scenarios with severe occlusions.}
    \label{fig:exp_sub_three_persons}     \vspace{-0.58cm}
\end{figure}

\begin{figure}[t]
\vspace{-0.05cm}
\begin{center}
\includegraphics[width=0.88\linewidth]{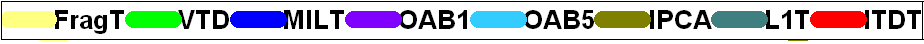}
   \includegraphics[width=0.88\linewidth]{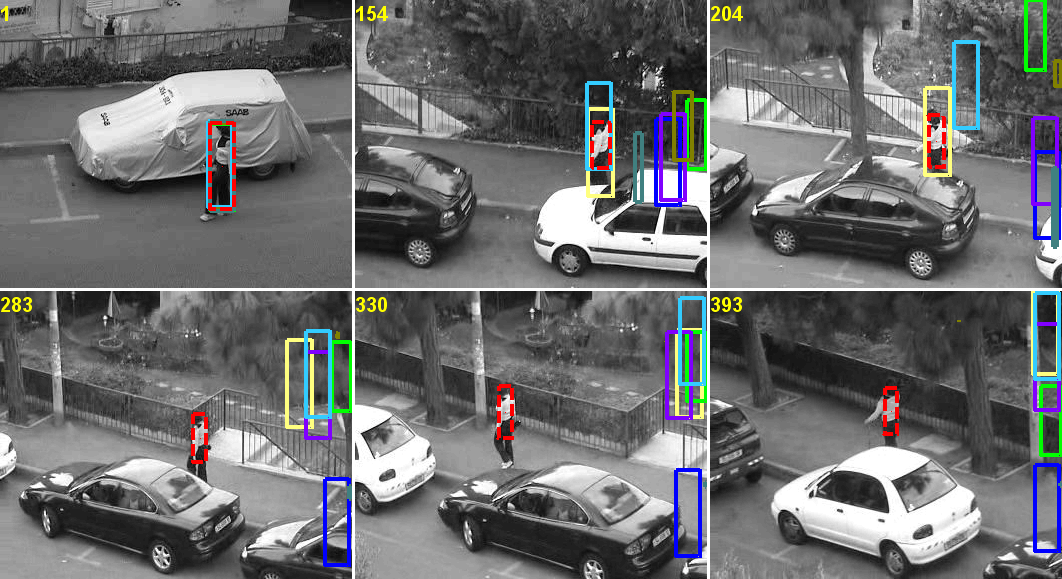}
\end{center}
\vspace{-0.8cm}
   \caption{The tracking results of the eight trackers over the
   representative frames (i.e., the 1st, 154th, 204th, 283rd, 330th, and 393rd frames)
   of the ``\textit{woman}'' video sequence
   in the
   scenarios with partial occlusions and body pose variations.}
    \label{fig:exp_woman}     \vspace{-0.8cm}
\end{figure}

\begin{figure}[t]
\vspace{-0.05cm}
\begin{center}
\includegraphics[width=0.88\linewidth]{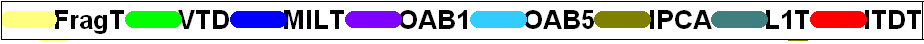}
   \includegraphics[width=0.88\linewidth]{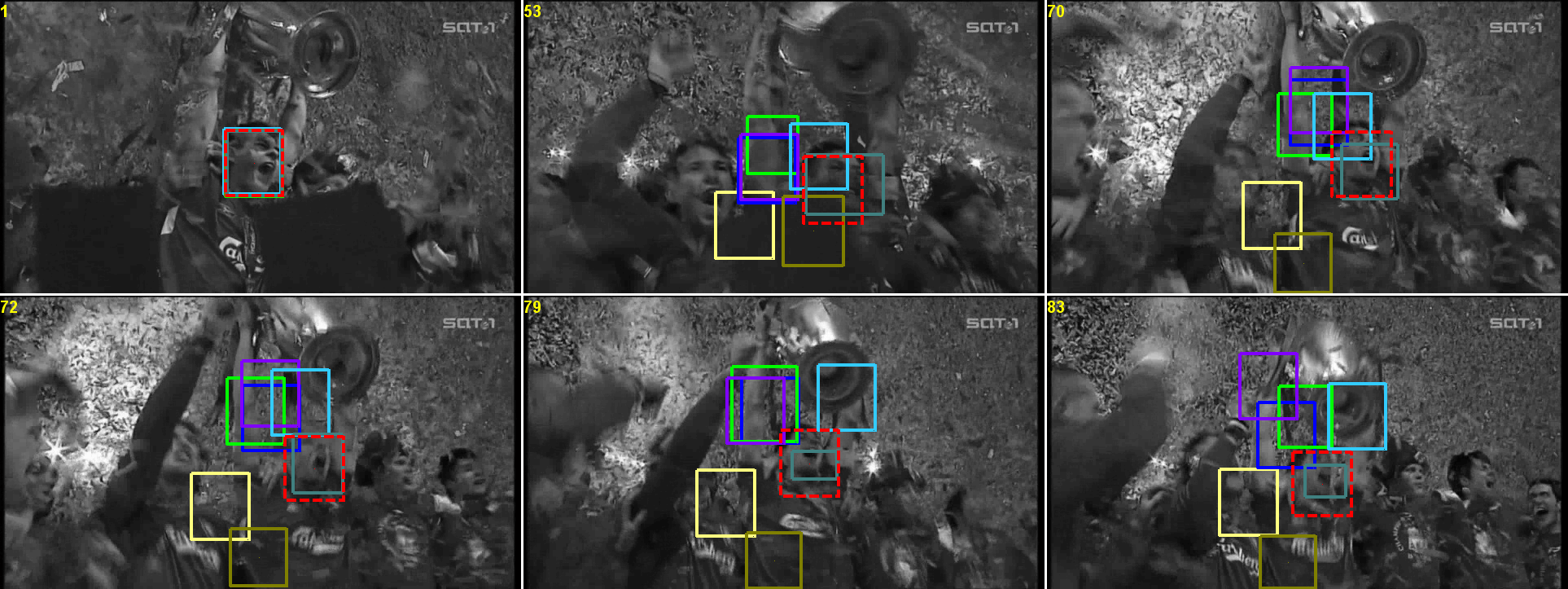}
\end{center}
\vspace{-0.8cm}
   \caption{The tracking results of the eight trackers over the
   representative frames (i.e., the 1st, 53rd, 70th, 72nd, 79th, and 83rd frames)
   of the ``\textit{soccer}'' video sequence
   in the
   scenarios with partial occlusions, head pose variations, background clutters, and motion blurring.}
    \label{fig:exp_soccer}     \vspace{-0.58cm}
\end{figure}

\begin{figure}[t]
\vspace{-0.05cm}
\begin{center}
\includegraphics[width=0.88\linewidth]{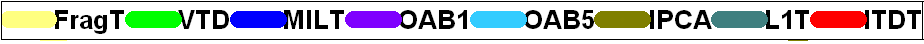}
   \includegraphics[width=0.88\linewidth]{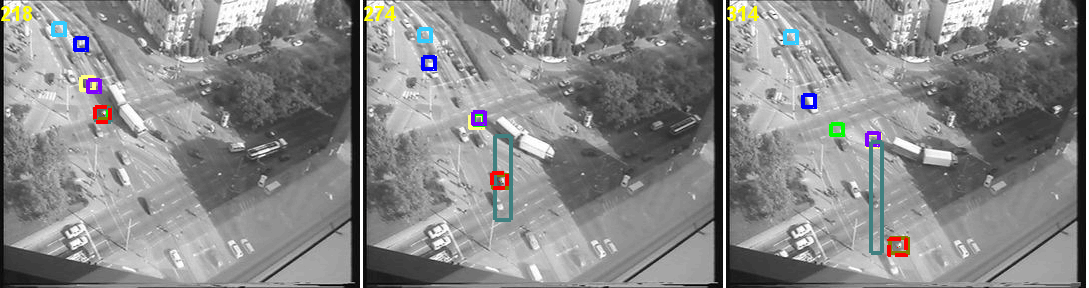}
\end{center}
\vspace{-0.8cm}
   \caption{The tracking results of the eight trackers over the
   representative frames (i.e., the 218th, 274th, and 314th frames)
   of the ``\textit{video-car}'' video sequence
   in the
   scenarios with small target and background clutter.}
    \label{fig:exp_video_bus2}     \vspace{-0.6cm}
\end{figure}

\begin{figure}[t]
\vspace{-0.05cm}
\begin{center}
\includegraphics[width=0.88\linewidth]{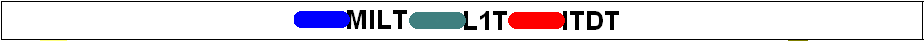}
   \includegraphics[width=0.88\linewidth]{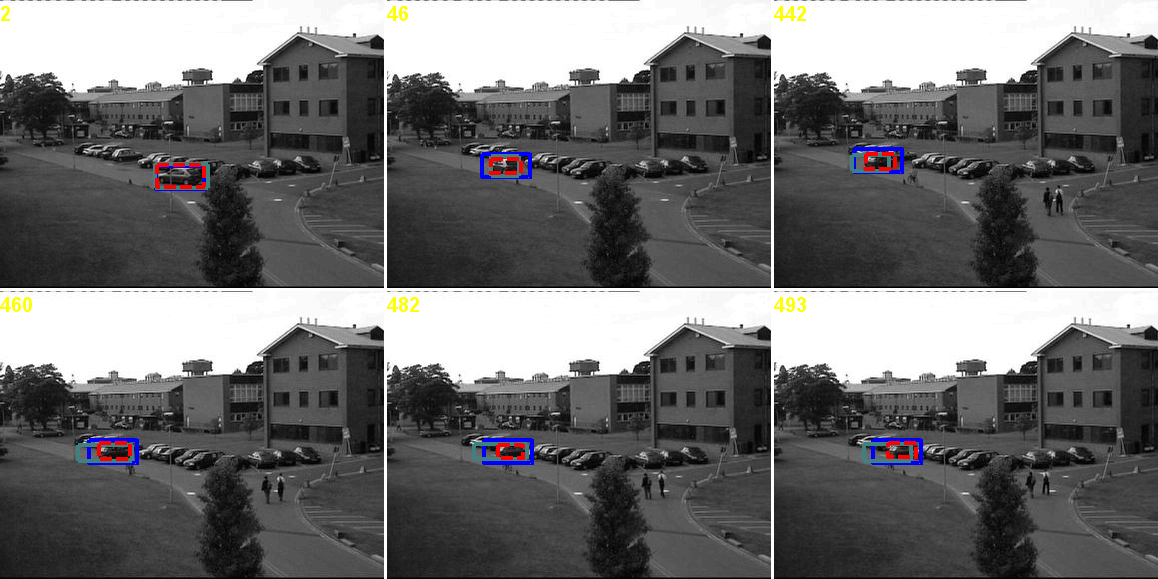}
\end{center}
\vspace{-0.8cm}
   \caption{The tracking results of the three best trackers (i.e., ITDT, MILT, and L1T for a better visualization) over the
   representative frames (i.e., the 2nd, 46th, 442nd, 460th, 482nd, and 493rd frames)
   of the ``\textit{pets-car}'' video sequence
   in the
   scenarios with partial occlusion and car pose variation.}
    \label{fig:exp_pets_car} \vspace{-0.5cm}
\end{figure}

\subsection{Tracking results}

\begin{figure}[t]
\vspace{-0.05cm}
\begin{center}
\includegraphics[width=0.88\linewidth]{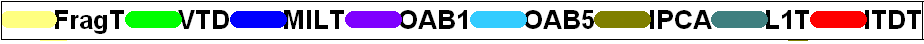}
   \includegraphics[width=0.88\linewidth]{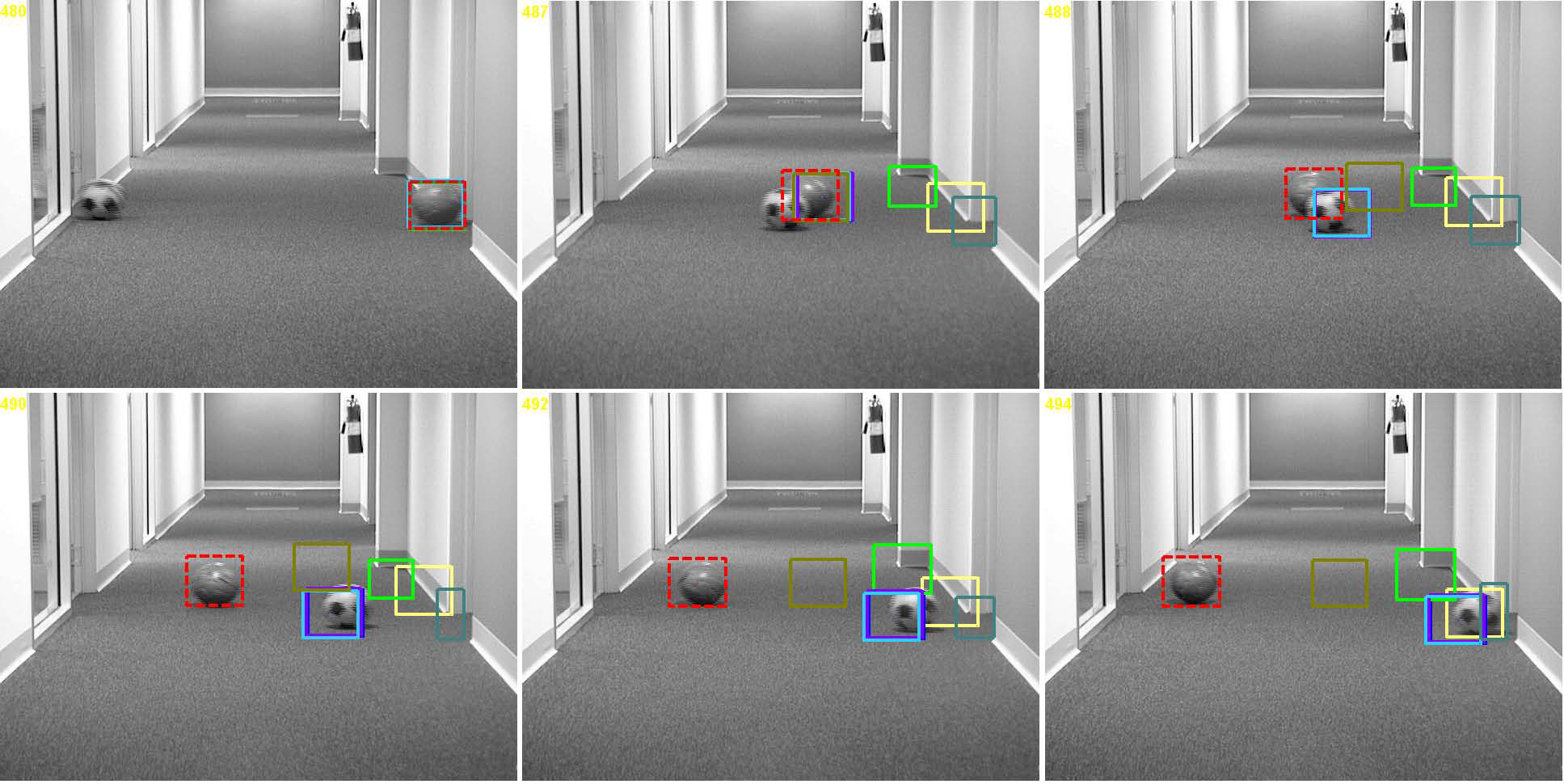}
\end{center}
\vspace{-0.8cm}
   \caption{The tracking results of the eight trackers over the
   representative frames (i.e., the 1st, 8th, 9th, 11th, 13th, and 15th)
   of the ``\textit{TwoBalls}'' video sequence in the
   scenarios with severe occlusions and motion blurring.}
    \label{fig:exp_two_balls} \vspace{-0.8cm}
\end{figure}

\begin{figure}[t]
\vspace{-0.05cm}
\begin{center}
\includegraphics[width=0.88\linewidth]{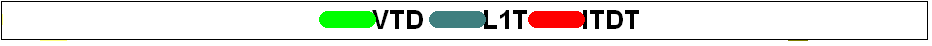}
   \includegraphics[width=0.88\linewidth]{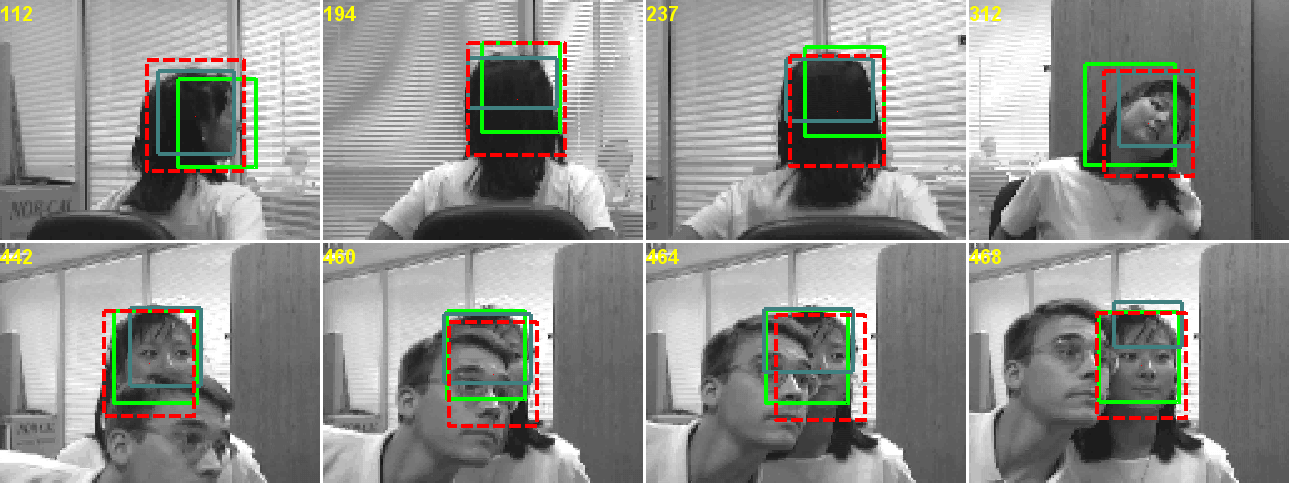}
\end{center}
\vspace{-0.8cm}
   \caption{The tracking results of the three best trackers (i.e., ITDT, L1T, and VTD for a better visualization) over the
   representative frames (i.e., the 112th, 194th, 237th, 312th, 442nd, 460th, 464th, and 468th frames)
   of the ``\textit{girl}'' video sequence
   in the
   scenarios with severe occlusion, in-plane/out-of-plane rotation, and head pose variation.}
    \label{fig:exp_girl}     \vspace{-0.7cm}
\end{figure}

\begin{figure}[t]
\vspace{-0.05cm}
\hspace{-0.9cm}
\begin{center}
\includegraphics[width=0.88\linewidth]{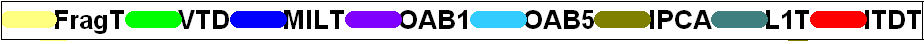}
   \includegraphics[width=0.88\linewidth]{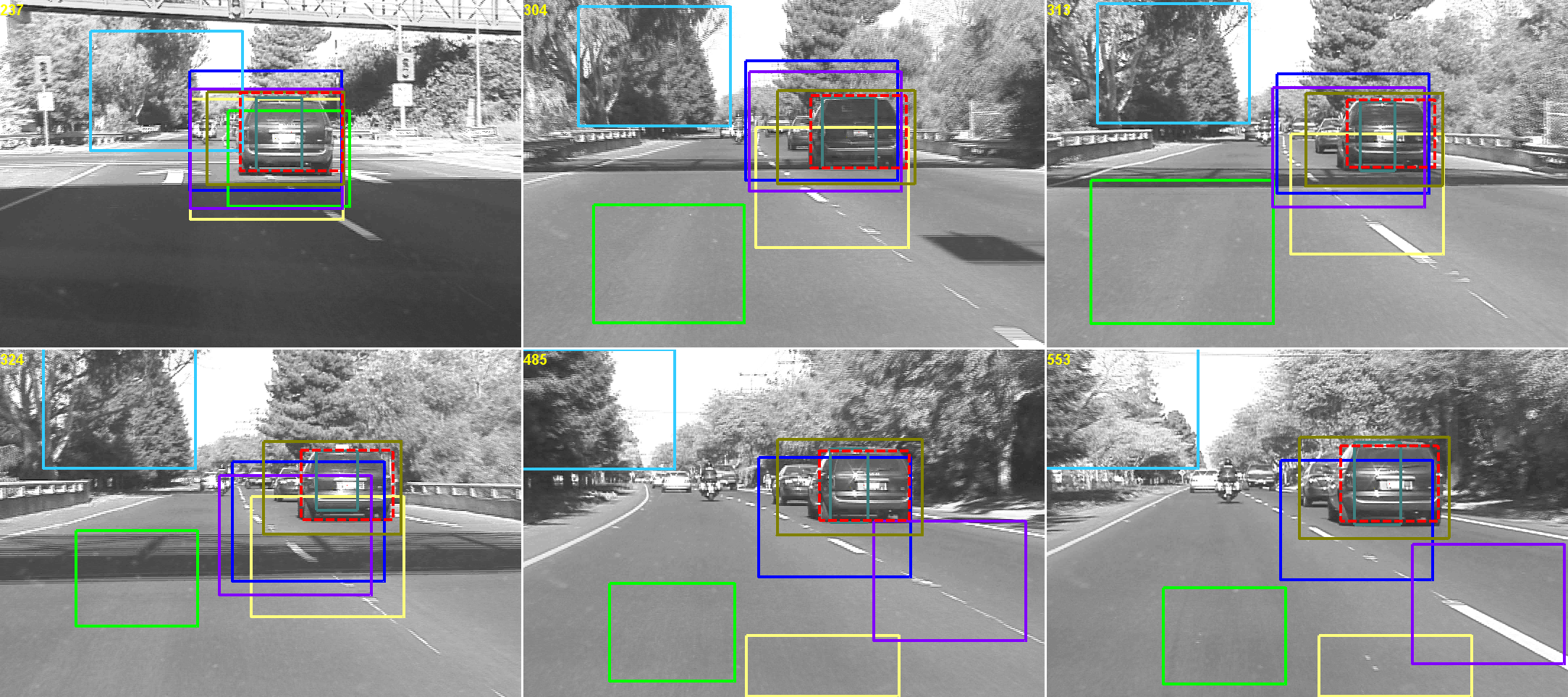}
\end{center}
\vspace{-0.8cm}
   \caption{The tracking results of the eight trackers over the
   representative frames (i.e., the 237th, 304th, 313th, 324th, 485th, and 553rd frames)
   of the ``\textit{car4}'' video sequence
   in the
   scenarios with shadow disturbance and pose variation.}
    \label{fig:exp_car4}     \vspace{-0.8cm}
\end{figure}

Due to space limit, we only report
tracking results for the eight trackers (highlighted by the bounding boxes in different colors)  over
representative frames of the first twelve video sequences, as shown in Figs.~\ref{fig:exp_trellis70}--\ref{fig:exp_car4}
(the caption of each figure includes the name of its corresponding video sequence).
Complete quantitative comparisons for all the twenty video sequences can be found in
Tab.~\ref{Tab:quantitative}.

As shown in
Fig.~\ref{fig:exp_trellis70}, a man walks under a treillage\footnote[7]{Downloaded from http://www.cs.toronto.edu/$\sim$dross/ivt/.}.
Suffering from large changes in
environmental illumination and head pose,  VTD and OAB5 start to fail in tracking the face after
the 170th frame while OAB1, IPCA, MILT, and FragT  break down after the 182nd, 201st, 202nd, and 205th frames,  respectively.
L1T fails to track the face from the 252nd frame.
In contrast to these competing
trackers, the proposed ITDT is able to successfully track the face till the end of the video.

Fig.~\ref{fig:exp_tiger} shows that a tiger toy is shaken strongly\footnote[8]{Downloaded from http://vision.ucsd.edu/$\sim$bbabenko/project$\_$miltrack.shtml.}.
Affected by drastic pose variation, illumination change, and partial occlusion,
L1T, IPCA, OAB5, and FragT fail in tracking the tiger toy after the
72nd, 114th, 154th, and 224th frames, respectively.
From the 113th frame,
VTD fails to track the tiger toy intermittently.
OAB1 is not lost in tracking the tiger toy, but it achieves inaccurate tracking
results.
In contrast, both MILT and ITDT are capable of accurately tracking the tiger toy in the situations of
illumination changes and partial occlusions.

As shown in Fig.~\ref{fig:exp_car11},  there is a car moving quickly in a dark road scene with
background clutter and varying lighting conditions\footnote[9]{Downloaded from http://www.cs.toronto.edu/$\sim$dross/ivt/.}.
After the 271st frame, VTD fails to track the car due to illumination changes. Distracted by background clutter,
MILT, FragT, L1T, and OAB1 break down after the 196th, 208th, 286th, and 295th frames, respectively.
OAB5 can keep tracking the car, but obtain inaccurate tracking results.
In contrast, only ITDT and IPCA succeed in accurately tracking the car throughout the video sequence.

Fig.~\ref{fig:exp_animal} shows that several deer run and jump in a river\footnote[10]{Downloaded from http://cv.snu.ac.kr/research/$\sim$vtd/.}.
Because of drastic pose variation and motion blurring,
FragT fails in tracking the head of a deer after the 5th frame while IPCA, VTD, OAB1, and OAB5 lose the head of the deer
after the 13th, 17th, 39th, and 52nd frames, respectively. L1T and MILT are incapable of accurately tracking the head of the deer all the time,
and lose the target intermittently. Compared with these trackers, the proposed ITDT
is able to accurately track the head of the deer throughout the video sequence.

In the video sequence shown in Fig.~\ref{fig:exp_sub_three_persons},  several persons walk along a corridor\footnote[11]{Downloaded from http://homepages.inf.ed.ac.uk/rbf/caviardata1/.}. One person is
occluded severely by the other two persons.
All the competing trackers except for FragT and ITDT
suffer from
severe occlusion taking place between the 56th frame and the 76th frame. As a result, they fail to track the person after the 76th frame thoroughly.
On the contrary, FragT and ITDT can track the person successfully. However, FragT achieves less accurate tracking results
than ITDT.

Fig.~\ref{fig:exp_woman} shows that woman with varying body poses walks along a pavement\footnote[12]{Downloaded from http://www.cs.technion.ac.il/$\sim$amita/fragtrack/fragtrack.htm.}. In the meantime,
her body is occluded by several cars.
After the 127th frame, MILT, OAB1, IPCA, and VTD
start to drift away from the woman as a result of partial occlusion. L1T begins to lose the woman after the 147th frame while OAB5 fails to
track the woman from the 205th frame. From the 227th frame, FragT  stays far away from the woman.
Only ITDT can keep tracking the woman over time.

In the video sequence shown in Fig.~\ref{fig:exp_soccer},
a number of soccer players assemble together and scream excitedly, jumping up and down\footnote[13]{Downloaded from http://cv.snu.ac.kr/research/$\sim$vtd/.}. Moreover, their heads are partially occluded by many pieces of floating
paper.  FragT, IPCA, MILT, and OAB5 fail to track the face from the  49th, 52nd, 49th, and 87th frames, respectively.
From the 48th frame to the 94th frame, VTD and OAB1 achieve unsuccessful tracking performances.
After the 94th frame, they capture the location of the face again. Compared with these competing trackers,
the proposed ITDT can achieve good performance throughout the video sequence.

In Fig.~\ref{fig:exp_video_bus2}, several small-sized cars densely surrounded
by other cars move in a blurry traffic scene\footnote[14]{Downloaded from  http://i21www.ira.uka.de/image$\_$sequences/.}.
Due to the influence of background distraction
and small target, MILT, OAB5, FragT, OAB1, VTD, and L1T fail to track the car from the 69th, 160th, 190th, 196th,
246th, and 314th frames, respectively.
In contrast, both ITDT and IPCA are able to locate the car accurately at all times.

As shown in Fig.~\ref{fig:exp_pets_car}, a driver tries to parallel park in the gap between two cars\footnote[15]{Downloaded from  http://www.hitech-projects.com/euprojects/cantata/datasets$\_$cantata/dataset.html.}.
At the end of the video sequence, the car is
partially occluded by another car.  FragT, VTD, OAB1, and IPCA
achieve inaccurate tracking performances after the 122nd frame.  Subsequently, they begin to drift away after
the 435th frame, while OAB5 begins to break down from the 486th frame.
MILT and L1T are able to track the car, but achieve inaccurate tracking results.
In contrast to these competing trackers,
the proposed ITDT is able to perform accurate car tracking throughout the video.

Fig.~\ref{fig:exp_two_balls} shows that two balls are rolled
on the floor. In the middle of the video sequence, one ball is occluded by the other ball.
L1T, FragT  and VTD fail in tracking the ball in the 3rd, 5th, and 6th frames, respectively.
Before the 8th frame, OAB1, OAB5, MILT, and IPCA achieves inaccurate tracking results. After that,
IPCA fails to track the ball thoroughly while OAB1, OAB5, and MILT are distracted by another ball due to severe
occlusion. In contrast, only ITDT can successfully track the ball continuously even in the case of
severe occlusion.

In the video sequence shown in Fig.~\ref{fig:exp_girl}, a girl rotates her body drastically\footnote[16]{Downloaded from  http://vision.ucsd.edu/$\sim$bbabenko/project$\_$miltrack.shtml.}.
At the end, her face is occluded by the other person's face.
Suffering from severe occlusion, IPCA fails to track the face from the 442nd frame while OAB5
begins to break down after the 486th frame. Due to the influence of the head's out-of-plane rotation,
MILT, OAB1, OAB5, FragT, and L1T obtain inaccurate tracking results from the 88th frame to the 265th frame.
VTD can track the face persistently, but achieves inaccurate tracking results
in most frames.
On the contrary, the proposed ITDT can achieve accurate tracking results throughout the video sequence.

As shown in Fig.~\ref{fig:exp_car4}, a car is moving in a highway\footnote[17]{Downloaded from  http://www.cs.toronto.edu/$\sim$dross/ivt/.}.
Due to the influence of both shadow disturbance and pose variation,
OAB5 and OAB1 fail to track the car thoroughly after the 241st and 331st frames, respectively.
In contrast, VTD is able to track the car before the 240th frame. However, it
tracks the car inaccurately or unsuccessfully after the 240th frame.
MILT begin to achieve inaccurate tracking results after the 323rd frame.
In contrast, ITDT can track the car accurately in the situations of shadow disturbance and pose
variation throughout the video sequence,
while both IPCA and L1T achieve less accurate tracking results than ITDT.

\subsection{Quantitative comparison}
\label{exp:quantitative_comparison}

\begin{figure}[h!]
\begin{center}
 \includegraphics[width=.8\linewidth]{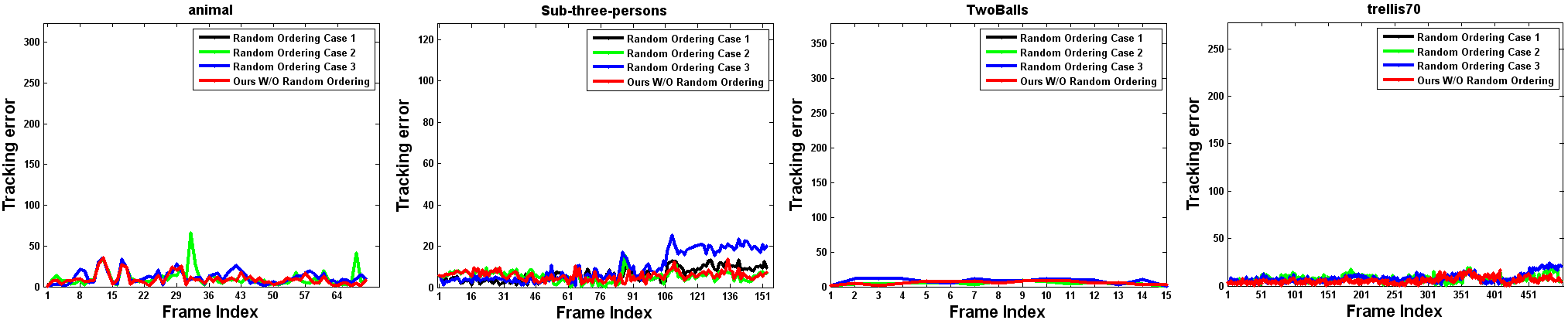}\\
\end{center}
\vspace{-0.8cm}
   \caption{Quantitative tracking performances using different cases of ``temporal ordering'' (obtained by small-scale random permutation)
   on the four video sequences. The error curves of the four video sequences in this figure have
   the same y-axis scale as those of the four video sequences in Fig.~\ref{fig:exp_error_curve}.}
    \label{fig:temporalorderingexperiments} \vspace{-0.8cm}
\end{figure}

\subsubsection{Evaluation criteria}

For all the twenty video sequences, the object center locations are labeled manually and used as the ground
truth. Hence, we can quantitatively evaluate the
performances of the eight trackers  by computing their pixel-based
tracking location errors from the ground truth.

In order to better evaluate the quantitative tracking performance of each tracker, we define a
criterion called the tracking success rate ({TSR}) as:
$\mbox{TSR} = \frac{N_{s}}{N}$.  Here $N$ is the total number of the frames from a video sequence,
and $N_{s}$ is the number of the frames in which a tracker can successfully track the target. The larger the value of TSR is, the better performance the tracker achieves.
Furthermore, we introduce an evaluation criterion to determine the success or failure of
tracking in each frame: $\frac{\mbox{TLE}}{\mbox{max}\left(W, H\right)}$,
where \mbox{TLE} is the pixel-based
tracking location error with respect to the ground truth, $W$ is the width of the ground truth bounding box for object localization,
and  $H$ is the height of the ground truth bounding box.
If $\frac{\mbox{TLE}}{\mbox{max}\left(W, H\right)}<0.25$, the tracker is considered to be successful;
otherwise, the tracker fails.
For each tracker, we compute its corresponding TSRs for all the video sequences.
These TSRs are finally used as the criterion for the quantitative evaluation of each tracker.

\begin{figure}[h]
\begin{center}
\includegraphics[width=.8\linewidth]{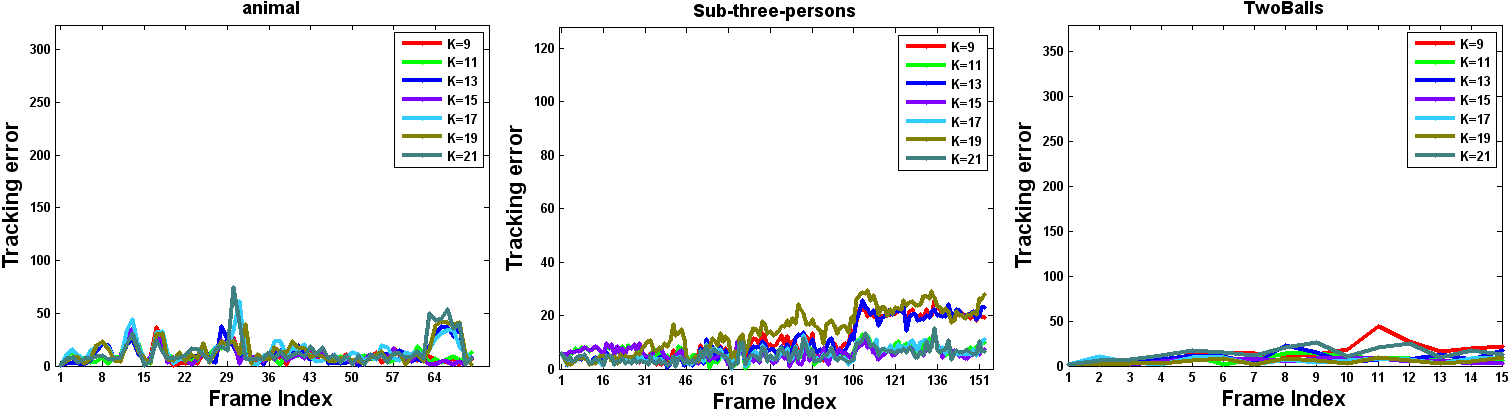}\\
\end{center}
\vspace{-0.8cm}
   \caption{Quantitative tracking performances using different choices of $K$ on the three video sequences.
    The error curves of the three video sequences in this figure have
   the same y-axis scale as those of the three video sequences in Fig.~\ref{fig:exp_error_curve}.} \vspace{-0.1cm}
   \label{fig:NNSelection} \vspace{-0.6cm}
\end{figure}

\subsubsection{Investigation of nearest neighbor construction}
\label{exp:quantitative_comparison_nn}
The $K$ nearest neighbors used in our 3D-DCT representation are
always ordered according to their distances to the current sample (as described in Sec.~\ref{sec:likelihood_eval}).
In order to examine the influence of sorting such $K$ nearest neighbors,
we randomly exchange a few of them and perform the tracking experiments again,
as shown in Fig.~\ref{fig:temporalorderingexperiments}.
It is seen from Fig.~\ref{fig:temporalorderingexperiments} that
the tracking performances using different ordering cases
are close to each other.

In order to evaluate the effect of nearest neighbor selection, we
conduct one experiment on three video sequences using
difference choices of $K$ such that $K \in \{9, 11, 13, 15, 17, 19, 21\}$, as
shown in Fig.~\ref{fig:NNSelection}.
From Fig.~\ref{fig:NNSelection}, we can see that the tracking performances using
different configurations of $K$ within a certain range are close to each other. Therefore, our
3D-DCT representation is not very sensitive to the choice of $K$
which lies in a certain interval.

\begin{figure}[t]
\begin{center}
   \includegraphics[width=0.3\linewidth]{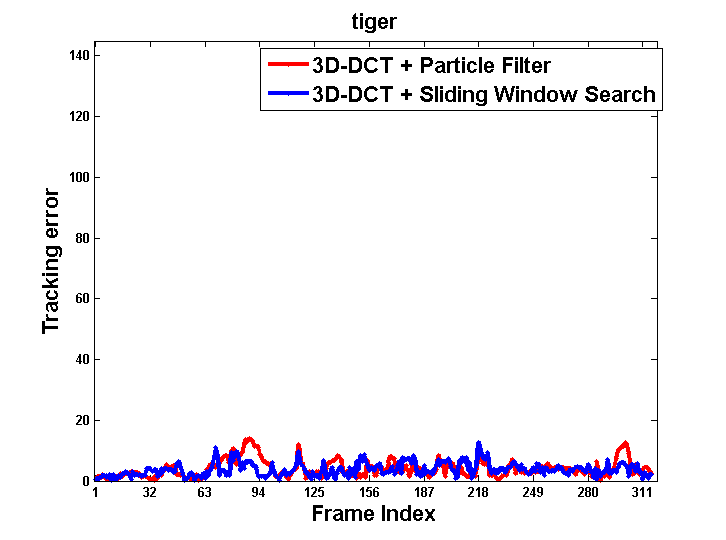} \hspace{-0.56cm}
   \includegraphics[width=0.3\linewidth]{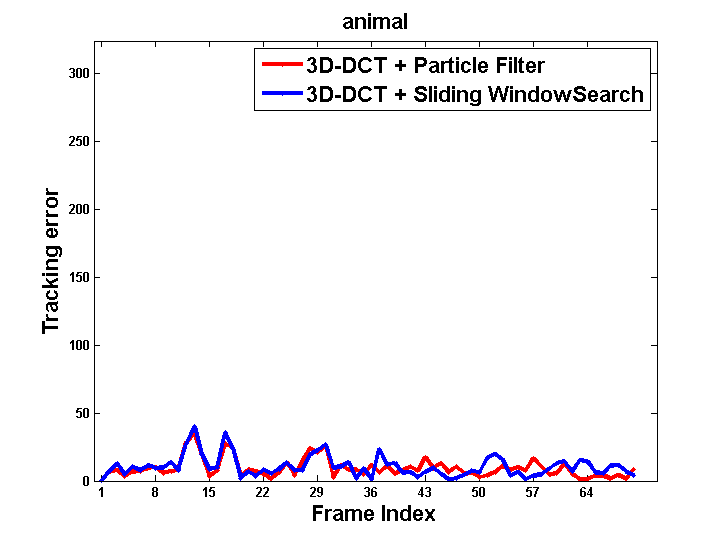} \hspace{-0.56cm}
   \includegraphics[width=0.3\linewidth]{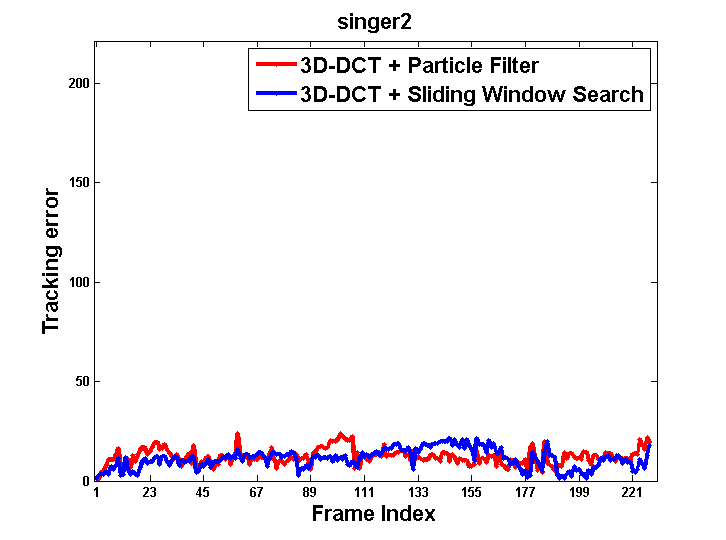} \hspace{-0.56cm}
\end{center}
\vspace{-0.8cm}
   \caption{Quantitative tracking performances of different state inference methods, i.e., sliding window search-based object tracking (referred to as ``3D-DCT+Sliding Window Search'') and its comparison
   with particle filter-based tracking (referred to as ``3D-DCT + Particle Filter'') on the three video sequences.
    The error curves of the three video sequences in this figure have
   the same y-axis scale as those of the three video sequences in Fig.~\ref{fig:exp_error_curve} and the supplementary file.
   Clearly, their tracking performances are almost consistent with each other.
   }
    \label{fig:comparison_pf_no_pf} \vspace{-0.8cm}
\end{figure}

\subsubsection{Comparison of object representation and state inference}
\label{exp:state_inference}

From Tab.~\ref{Tab:quantitative}, we see that
our tracker achieves equal or higher tracking accuracies than the competing trackers in most cases.
Moreover, our tracker utilizes
the same state inference method (i.e., particle filter)
as IPCA, L1T, and VTD. Consequently, our 3D-DCT object representation
play a more critical role in improving the tracking performance
than those of IPCA, L1T, and VTD.

Furthermore, we make a performance comparison between our particle filter-based method (referred to as ``3D-DCT + Particle Filter'') and
a simple state inference method (referred to as ``3D-DCT + Sliding Window Search'').
Clearly, Fig.~\ref{fig:comparison_pf_no_pf} shows that
the tracking performances of two state inference methods are close to each other.
Besides, Tab.~\ref{Tab:quantitative} shows that our ``3D-DCT + Particle Filter'' obtains more accurate tracking results
than those of MILT and OAB, which also use a sliding window for state inference. Therefore, we conclude that the 3D-DCT object representation
is mostly responsible for the enhanced tracking performance relative to MILT and OAB.

\begin{figure}[t!]
\vspace{-0.08cm}
\begin{center}
\includegraphics[width=0.8\linewidth]{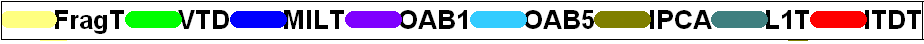}
   \includegraphics[width=0.86\linewidth]{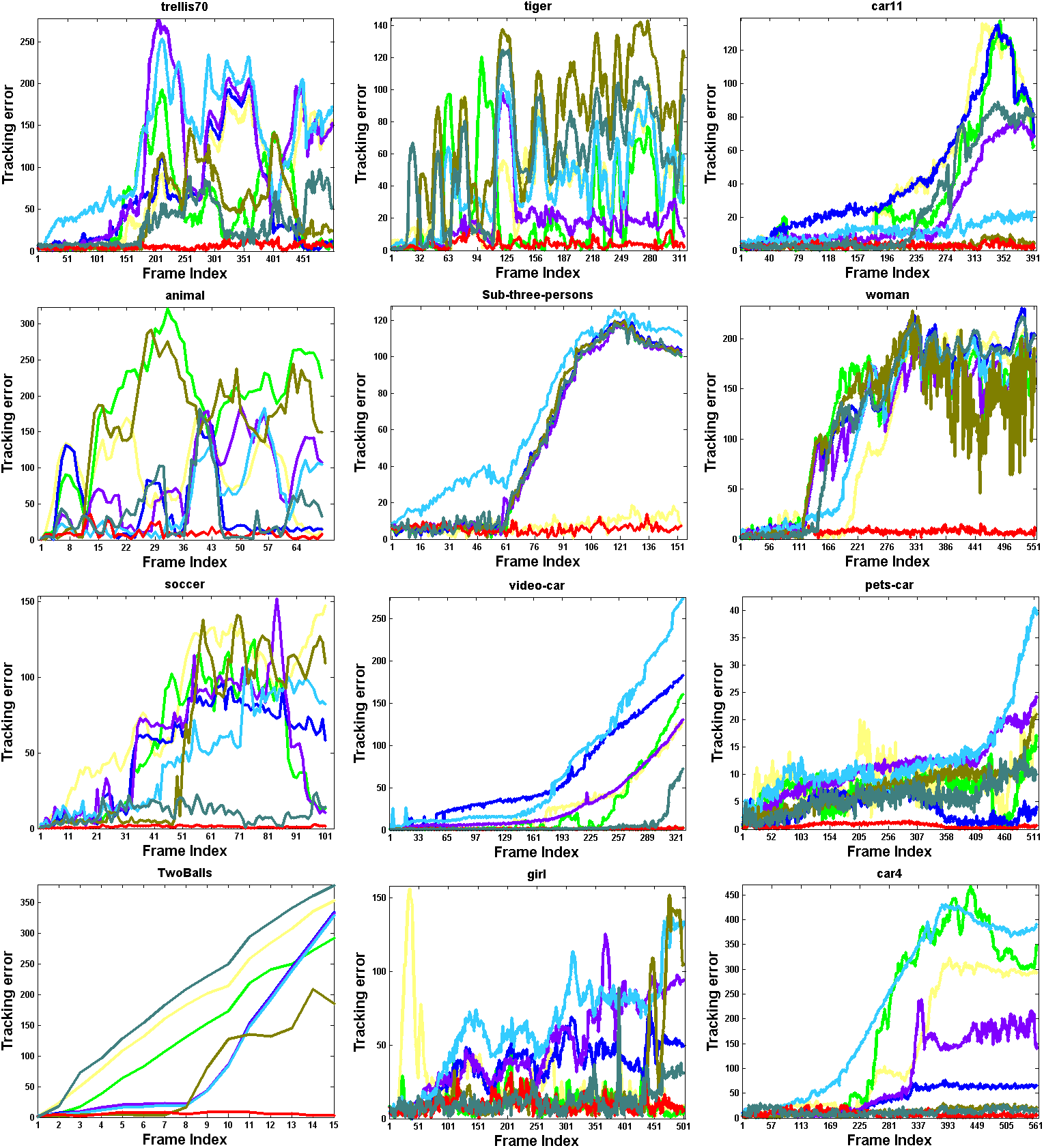}
\end{center}
\vspace{-0.8cm}
   \caption{The tracking location error plots obtained by the eight trackers over the first twelve videos. In each sub-figure, the x-axis corresponds to the frame index number, and the y-axis is associated with
the  tracking location error.}
    \label{fig:exp_error_curve}
    \vspace{-0.8cm}
\end{figure}

\begin{figure}[t]
\vspace{-0.08cm}
\begin{center}
   \includegraphics[width=0.86\linewidth]{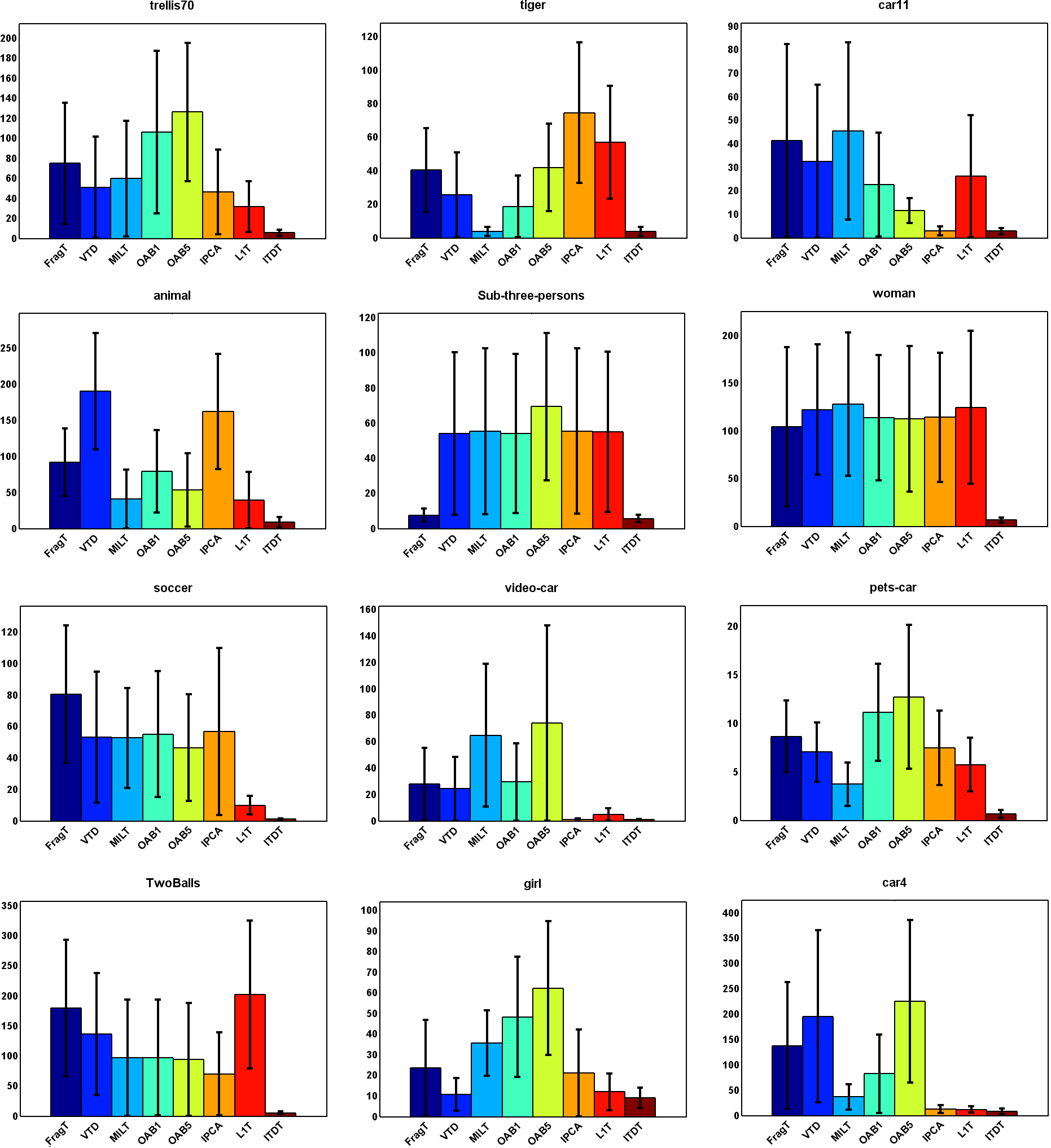}
\end{center}
\vspace{-0.8cm}
   \caption{The quantitative comparison results of the eight trackers over the first twelve videos.
The figure reports the mean and standard deviation of their tracking location
errors over the first twelve videos.
In each sub-figure, the x-axis shows the competing trackers, the y-axis is associated with
the means of their tracking location errors, and the error bars correspond to the standard deviations of their tracking location errors.
}
    \label{fig:exp_std_error_curve} \vspace{-0.7cm}
\end{figure}

\subsubsection{Comparison of competing trackers}
\label{sec:summary}

Fig.~\ref{fig:exp_error_curve} plots the tracking location errors (highlighted in different colors) obtained by the eight trackers
for the first twelve video sequences.
Furthermore, we also compute the mean and standard deviation of the tracking location errors for the first twelve video sequences,
and report the results in Fig.~\ref{fig:exp_std_error_curve}.

Moreover, Tab.~\ref{Tab:quantitative} reports all the corresponding TSRs of the eight trackers over the total twenty video
sequences.
From Tab.~\ref{Tab:quantitative}, we can see that
the mean and standard deviation of the TSRs obtained by the proposed ITDT
is respectively 0.9802 and 0.0449, which are the best among all the eight trackers.
The proposed ITDT also
achieves the largest TSR over 19 out of 20 video sequences.
As for the ``surfer'' video sequence, the proposed ITDT is slightly inferior to
the best MILT  (i.e., 1.33\% difference).
We believe this is because in the ``surfer'' video sequence, the tracked
object (i.e., the surfer's head) has an low-resolution appearance with
drastic motion blurring. In addition, the surfer's body has a similar color appearance
to the tracked object, which usually leads to the distraction of the trackers using color
information.  Furthermore,
the tracked object's appearance is varying greatly
due to the influence of pose variation and out-of-plane rotation.
Under such circumstances,
the trackers using local features are usually more effective than those using global features.
Therefore, the MILT using Haar-like features
slightly outperforms the proposed ITDT using color features in the ``surfer'' video sequence.
In summary, the 3D-DCT based object representation used by
the proposed ITDT
is able to
exploit the correlation between the current appearance sample
and the previous appearance samples in the 3D-DCT reconstruction process,
and encodes the discriminative information from object/non-object classes.
This may have contributed to the tracking robustness in complicated
scenarios (e.g., partial occlusions and pose variations).

\begin{table}
\caption{The quantitative comparison results of the eight trackers over the twenty video sequences.
The table reports their tracking success  rates (i.e., TSRs) over each video sequence.  \vspace{-0.3cm}
}
\label{Tab:quantitative}
\scriptsize
\centering
\begin{tabular}{ c||c|c|c|c|c|c|c|c }
\hline 
            & \makebox[1.1cm]{FragT} & \makebox[1.1cm]{VTD} & \makebox[1.1cm]{MILT} & \makebox[1.1cm]{OAB1} & \makebox[1.1cm]{OAB5}  & \makebox[1.1cm]{IPCA} & \makebox[1.1cm]{L1T} & \makebox[1.1cm]{\textbf{ITDT}}\\
\hline\hline

trellis70   & 0.2974  &  0.4072  &  0.3493  &  0.2295  &   0.0339  &  0.3593  &  0.3972   & \textbf{1.0000}\\\hline
tiger   & 0.1672  &  0.5205  &  \textbf{0.9495}  &  0.2808  &  0.1767   & 0.1104   & 0.1451   & \textbf{0.9495}\\\hline
car11   & 0.4020  &  0.4326  &  0.1043  &  0.3181  &  0.2799   & 0.9211   & 0.5700   & \textbf{0.9898}\\\hline
animal   & 0.1408  &  0.0845  &  0.6761  &  0.3099  &  0.5352   & 0.1690   & 0.5352   & \textbf{0.9859}\\\hline
sub-three-persons   & \textbf{1.0000}  &  0.4610  &  0.4481  &  0.4610  &  0.2662   & 0.4481   & 0.4481   & \textbf{1.0000}\\\hline
woman    & 0.2852 &   0.2004 &   0.2058  &  0.2148 &   0.1859  &  0.2148 &   0.2509  &   \textbf{0.9530}\\\hline

soccer   & 0.1078  &  0.3824  &  0.2941  &  0.3725   & 0.4118   & 0.4902   & 0.9510   & \textbf{1.0000}\\\hline
video-car   & 0.4711  &  0.6353  &  0.1550  &  0.4225   & 0.0578   & \textbf{1.0000}   & 0.9058   & \textbf{1.0000}\\\hline

pets-car   & 0.2959   & 0.4062  &  0.8801  &  0.1799   & 0.1199   & 0.4081   & 0.6983   & \textbf{1.0000}\\\hline
two-balls   & 0.1250  &  0.2500  &  0.3125  &  0.3125   & 0.3750   & 0.5625   & 0.1250   & \textbf{1.0000}\\\hline

girl    & 0.6335  &  0.9044  &  0.2211  &  0.1773   & 0.1633   & 0.8466   & 0.8845   & \textbf{0.9741}\\\hline
car4    & 0.4139  &  0.3783  &  0.4849  &  0.4547   & 0.2327   & 0.9982   & \textbf{1.0000}   & \textbf{1.0000}\\\hline
shaking   & 0.1534  &  0.2767  &  0.9918 &   0.9890   & 0.8438    &0.0110 &   0.0411 & \textbf{0.9973} \\ \hline

pktest02   & 0.1667  &  \textbf{1.0000}  &  \textbf{1.0000}  &  \textbf{1.0000}   & 0.2333   & \textbf{1.0000}   & \textbf{1.0000}   & \textbf{1.0000}\\\hline
davidin300   &  0.4545 &   0.7900  &  0.9654  &  0.3550  &  0.4762 &   \textbf{1.0000}  &  0.8528   & \textbf{1.0000}\\ \hline

surfer   & 0.2128  &  0.4149  &  \textbf{0.9894}  &  0.3112   & 0.0399   & 0.4069   & 0.2766   & 0.9761\\\hline
singer2   & 0.9304 &   \textbf{1.0000}  &  \textbf{1.0000} &   0.3783  &  0.2087  &  \textbf{1.0000} &   0.6739 & \textbf{1.0000}\\\hline

seq-jd   & \textbf{0.8020}  &  0.7723 &   0.5545   & 0.5446    &0.3168   & 0.6634  &  0.2277 &   \textbf{0.8020}\\\hline

cubicle   & 0.7255  &  0.9020  &  0.2353  &  0.4706   & 0.8627   & 0.7255   & 0.6863   & \textbf{1.0000}\\\hline
seq-simultaneous   & 0.6829  &  0.3171  &  0.2927 &   0.6829 &   0.6585  &  0.3171   & 0.5854    & \textbf{0.9756} \\

\hline
\hline
mean &     0.4234   & 0.5268 &   0.5555    &0.4233   & 0.3239   & 0.5826    &0.5629   & \textbf{0.9802} \\ \hline
s.t.d. &     0.2817 &   0.2768 &   0.3382&    0.2315   & 0.2438  &  0.3360&    0.3126 &   \textbf{0.0449}\\
\hline
\end{tabular}
\end{table}

\vspace{-0.28cm}
\section{Conclusion \label{sec:conclusion}}
\vspace{-0.16cm}

In this paper, we have proposed an effective tracking algorithm based on the 3D-DCT.
In this algorithm, a compact object representation has been constructed
using the 3D-DCT, which can produce a compact energy spectrum whose high-frequency
components are discarded. The problem of constructing the compact object representation has been converted to that of
how to efficiently compress and reconstruct the video data. To efficiently update the object representation during tracking,
we have also proposed an incremental 3D-DCT algorithm which decomposes the 3D-DCT into the successive operations of the 2D-DCT and 1D-DCT
on the video data. The incremental 3D-DCT algorithm only needs to compute 2D-DCT for newly added frames as well as the 1D-DCT along the time dimension, leading to
high computational efficiency. Moreover, by computing and storing the cosine basis functions beforehand, we can
significantly reduce the computational complexity of the 3D-DCT. Based on the incremental 3D-DCT algorithm, a discriminative criterion
has been designed to measure the information loss resulting from 3D-DCT based signal reconstruction,
which contributes to
evaluating the confidence score of a test sample belonging to the foreground object.
Since considering
both the foreground and the background reconstruction information, the discriminative criterion
is robust to complicated appearance changes (e.g., out-of-plane rotation and partial occlusion).
Using this discriminative criterion, we have conducted visual tracking in the particle filtering framework which propagates sample distributions over time.
Compared with several state-of-the-art trackers on  challenging video sequences, the proposed  tracker is  more
robust to the challenges including illumination changes, pose variations, partial occlusions, background distractions, motion blurring, complicated appearance changes, etc.
Experimental results
have demonstrated the effectiveness and robustness of the proposed  tracker.
\end{spacing}

\section*{Acknowledgments}

    This work is supported by ARC Discovery Project (DP1094764).

    All correspondence should be addressed to X. Li. 

{

\bibliographystyle{ieee-cs}

\begin{thebibliography}{10}

\bibitem{Limy-Ross17}
D.~A. Ross, J.~Lim, R.~Lin, and M.~Yang,
\newblock ``Incremental learning for robust visual tracking,''
\newblock {\em Int. J. Computer Vision}, vol. 77, no. 1, pp. 125--141, 2008.

\bibitem{lixi-cvpr2008}
X.~Li, W.~Hu, Z.~Zhang, X.~Zhang, M.~Zhu, and J.~Cheng,
\newblock ``Visual tracking via incremental log-euclidean riemannian subspace
  learning,''
\newblock in {\em Proc. IEEE Conf. Computer Vision \& Pattern Recognition},
  2008, pp. 1--8.

\bibitem{AKJ-DCT}
A.~K. Jain,
\newblock {\em Fundamentals of Digital Image Processing},
\newblock New Jersey: Prentice Hall Inc., 1989.

\bibitem{khayam-tr2003}
S.~A. Khayam,
\newblock ``The discrete cosine transform ({DCT}): theory and application,''
\newblock {\em Technical report, \emph{Michigan State University}}, 2003.

\bibitem{HAFED-LEVINE-IJCV2001}
Z.~M. Hafed and M.~D. Levine,
\newblock ``Face recognition using the discrete cosine transform,''
\newblock {\em Int. J. Computer Vision}, vol. 43, no. 3, pp. 167--188, 2001.

\bibitem{Feng-PR2003}
G.~Feng and J.~Jiang,
\newblock ``{JPEG} compressed image retrieval via statistical features,''
\newblock {\em Pattern Recognition}, vol. 36, no. 4, pp. 977--985, 2003.

\bibitem{He-ICIP2009}
D.~He, Z.~Gu, and N.~Cercone,
\newblock ``Efficient image retrieval in dct domain using hypothesis testing,''
\newblock in {\em Proc. Int. Conf. Image Processing}, 2009, pp. 225--228.

\bibitem{Chen-Liu-Sun-Yang-TMM2008}
D.~Chen, Q.~Liu, M.~Sun, and J.~Yang,
\newblock ``Mining appearance models directly from compressed video,''
\newblock {\em {IEEE} Trans. Multimedia}, vol. 10, no. 2, pp. 268--276, 2008.

\bibitem{Zhong-Zhang-Jain-TPAMI2000}
Y.~Zhong, H.~Zhang, and A.~K. Jain,
\newblock ``Automatic caption localization in compressed video,''
\newblock {\em {IEEE} Trans. Pattern Analysis \& Machine Intelligence}, vol.
  22, no. 4, pp. 385--392, 2000.

\bibitem{Adam-Fragment-2006}
A.~Adam, E.~Rivlin, and I.~Shimshoni,
\newblock ``Robust fragments-based tracking using the integral histogram,''
\newblock in {\em Proc. IEEE Conf. Computer Vision \& Pattern Recognition},
  2006, pp. 798--805.

\bibitem{Shen-Kim-Wang-TCSVT2010}
C.~Shen, J.~Kim, and H.~Wang,
\newblock ``Generalized kernel-based visual tracking,''
\newblock {\em {IEEE} Trans. Circuits \& Systems for Video Technology}, vol.
  20, no. 1, pp. 119--130, 2010.

\bibitem{Wang-Suter-Schindler-PAMI2007}
H.~Wang, D.~Suter, K.~Schindler, and C.~Shen,
\newblock ``Adaptive object tracking based on an effective appearance filter,''
\newblock {\em {IEEE} Trans. Pattern Analysis \& Machine Intelligence}, vol.
  29, no. 9, pp. 1661--1667, 2007.

\bibitem{Jepson-Fleet-Yacoob5}
A.~D. Jepson, D.~J. Fleet, and T.~F. El-Maraghi,
\newblock ``Robust online appearance models for visual tracking,''
\newblock in {\em Proc. IEEE Conf. Computer Vision \& Pattern Recognition},
  2001, pp. 415--422.

\bibitem{li2007robust}
X.~Li, W.~Hu, Z.~Zhang, X.~Zhang, and G.~Luo,
\newblock ``Robust visual tracking based on incremental tensor subspace
  learning,''
\newblock in {\em Proc. Int. Conf. Computer Vision}, 2007, pp. 1--8.

\bibitem{Meo-Ling-ICCV09}
X.~Mei and H.~Ling,
\newblock ``Robust visual tracking and vehicle classification via sparse
  representation,''
\newblock {\em {IEEE} Trans. Pattern Analysis \& Machine Intelligence}, 2011.

\bibitem{Liu-Yang-Huang-Meer-Gong-Kulikowski-eccv2010}
B.~Liu, L.~Yang, J.~Huang, P.~Meer, L.~Gong, and C.~Kulikowski,
\newblock ``Robust and fast collaborative tracking with two stage sparse
  optimization,''
\newblock in {\em Proc. Euro. Conf. Computer Vision}, 2010.

\bibitem{Liu-Huang-Kulikowski-Yang-cvpr2011}
B.~Liu, J.~Huang, C.~Kulikowski, and L.~Yang,
\newblock ``Robust tracking using local sparse appearance model and
  k-selection,''
\newblock in {\em Proc. IEEE Conf. Computer Vision \& Pattern Recognition},
  2011.

\bibitem{Li-Shen-Shi-cvpr2011}
H.~Li, C.~Shen, and Q.~Shi,
\newblock ``Real-time visual tracking with compressed sensing,''
\newblock in {\em Proc. IEEE Conf. Computer Vision \& Pattern Recognition},
  2011.

\bibitem{licvpr2012}
X.~Li, C.~Shen, Q.~Shi, D.~Anthony, and A.~van~den Hengel,
\newblock ``Non-sparse linear representations for visual tracking with online
  reservoir metric learning,''
\newblock {\em Proc. IEEE Conf. Computer Vision \& Pattern Recognition}, 2012.

\bibitem{Kwon-Lee-CVPR2010}
J.~Kwon and K.~M. Lee,
\newblock ``Visual tracking decomposition,''
\newblock in {\em Proc. IEEE Conf. Computer Vision \& Pattern Recognition},
  2010, pp. 1269--1276.

\bibitem{Porikli-Tuzel-Meer-CVPR2006}
F.~Porikli, O.~Tuzel, and P.~Meer,
\newblock ``Covariance tracking using model update based on lie algebra,''
\newblock in {\em Proc. IEEE Conf. Computer Vision \& Pattern Recognition},
  2006, pp. 728--735.

\bibitem{Wu-Cheng-Wang-Lu-iccv2009}
Y.~Wu, J.~Cheng, J.~Wang, and H.~Lu,
\newblock ``Real-time visual tracking via incremental covariance tensor
  learning,''
\newblock in {\em Proc. Int. Conf. Computer Vision}, 2009, pp. 1631--1638.

\bibitem{Comaniciu-Ramesh-Meer-TPAMI}
D.~Comaniciu, V.~Ramesh, and P.~Meer,
\newblock ``Kernel-based object tracking,''
\newblock {\em {IEEE} Trans. Pattern Analysis \& Machine Intelligence}, vol.
  25, no. 5, pp. 564--577, 2003.

\bibitem{Shen-Brooks-van-den-Hengel-TIP2007}
C.~Shen, M.~J. Brooks, and A.~van~den Hengel,
\newblock ``{Fast Global Kernel Density Mode Seeking}: {Applications To
  Localization And Tracking},''
\newblock {\em {IEEE} Trans. Image Processing}, vol. 16, no. 5, pp. 1457--1469,
  2007.

\bibitem{Qu-Schonfeld-TIP2008}
W.~Qu and D.~Schonfeld,
\newblock ``Robust control-based object tracking,''
\newblock {\em {IEEE} Trans. Image Processing}, vol. 17, no. 9, pp. 1721--1726,
  2008.

\bibitem{Avidan-2004}
S.~Avidan,
\newblock ``Support vector tracking,''
\newblock {\em {IEEE} Trans. Pattern Analysis \& Machine Intelligence}, vol.
  26, no. 8, pp. 1064--1072, 2004.

\bibitem{Tian-Zhang-Liu-ACCV2007}
M.~Tian, W.~Zhang, and F.~Liu,
\newblock ``On-line ensemble {SVM} for robust object tracking,''
\newblock in {\em Proc. Asian Conf. Computer Vision}, 2007, pp. 355--364.

\bibitem{Tang-Brennan-Tao-ICCV2007}
F.~Tang, S.~Brennan, Q.~Zhao, and H.~Tao,
\newblock ``Co-tracking using semi-supervised support vector machines,''
\newblock in {\em Proc. Int. Conf. Computer Vision}, 2007.

\bibitem{li2011graph}
X.~Li, A.~Dick, H.~Wang, C.~Shen, and A.~van~den Hengel,
\newblock ``Graph mode-based contextual kernels for robust svm tracking,''
\newblock in {\em Proc. Int. Conf. Computer Vision}, 2011, pp. 1156--1163.

\bibitem{Grabner-Grabner-Bischof-BMVC2006}
H.~Grabner, M.~Grabner, and H.~Bischof,
\newblock ``Real-time tracking via on-line boosting,''
\newblock in {\em Proc. British Machine Vision Conf.}, 2006, pp. 47--56.

\bibitem{Grabner-Grabner-Bischof-ECCV2008}
H.~Grabner, C.~Leistner, and H.~Bischof,
\newblock ``Semi-supervised on-line boosting for robust tracking,''
\newblock in {\em Proc. Euro. Conf. Computer Vision}, 2008, pp. 234--247.

\bibitem{Collins-Liu-Leordeanu-PAMI2005}
R.~T. Collins, Y.~Liu, and M.~Leordeanu,
\newblock ``Online selection of discriminative tracking features,''
\newblock {\em {IEEE} Trans. Pattern Analysis \& Machine Intelligence}, vol.
  27, no. 10, pp. 1631--1643, 2005.

\bibitem{Santner-Leistner-Saffari-Pock-Bischof-cvpr2010}
J.~Santner, C.~Leistner, A.~Saffari, T.~Pock, and H.~Bischof,
\newblock ``Prost: Parallel robust online simple tracking,''
\newblock in {\em Proc. IEEE Conf. Computer Vision \& Pattern Recognition},
  2010, pp. 723--730.

\bibitem{Babenko-Yang-Belongie-cvpr2009}
B.~Babenko, M.~Yang, and S.~Belongie,
\newblock ``Visual tracking with online multiple instance learning,''
\newblock in {\em Proc. IEEE Conf. Computer Vision \& Pattern Recognition},
  2009, pp. 983--990.

\bibitem{Fan-Wu-Dai-ECCV2010}
J.~Fan, Y.~Wu, and S.~Dai,
\newblock ``Discriminative spatial attention for robust tracking,''
\newblock in {\em Proc. Euro. Conf. Computer Vision}, 2010, pp. 480--493.

\bibitem{Wang-Hua-Han-eccv2010}
X.~Wang, G.~Hua, and T.~X. Han,
\newblock ``Discriminative tracking by metric learning,''
\newblock in {\em Proc. Euro. Conf. Computer Vision}, 2010, pp. 200--214.

\bibitem{Jiang-Liu-Wu-TIP2011}
N.~Jiang, W.~Liu, and Y.~Wu,
\newblock ``Learning adaptive metric for robust visual tracking,''
\newblock {\em {IEEE} Trans. Image Processing}, vol. 20, no. 8, pp. 2288--2300,
  2011.

\bibitem{Yang-Fan-Fan-Wu-TIP2009}
M.~Yang, Z.~Fan, J.~Fan, and Y.~Wu,
\newblock ``Tracking non-stationary visual appearances by data-driven
  adaptation,''
\newblock {\em {IEEE} Trans. Image Processing}, vol. 18, no. 7, pp. 1633--1644,
  2009.

\bibitem{Liu-Yu-ICCV2007}
X.~Liu and T.~Yu,
\newblock ``Gradient feature selection for online boosting,''
\newblock in {\em Proc. Int. Conf. Computer Vision}, 2007, pp. 1--8.

\bibitem{Avidan-2007}
S.~Avidan,
\newblock ``Ensemble tracking,''
\newblock {\em {IEEE} Trans. Pattern Analysis \& Machine Intelligence}, vol.
  29, no. 2, pp. 261--271, 2007.

\bibitem{Levy-Lindenbaum-JMAA2000}
L.~D. Lathauwer, B.~Moor, and J.~Vandewalle,
\newblock ``On the best rank-1 and rank-$(r_{1},r_{2},\ldots, r_{n})$
  approximation of higher-order tensors,''
\newblock {\em SIAM Journal of Matrix Analysis and Applications}, vol. 21, no.
  4, pp. 1324--1342, 2000.

\bibitem{Wang-Yang-Yu-Lv-Huang-Gong-CVPR2010}
J.~Wang, J.~Yang, K.~Yu, F.~Lv, T.~Huang, and Y.~Gong,
\newblock ``Locality-constrained linear coding for image classification,''
\newblock in {\em Proc. IEEE Conf. Computer Vision \& Pattern Recognition},
  2010, pp. 3360--3367.

\bibitem{Isard-Blake-ECCV1996}
M.~Isard and A.~Blake,
\newblock ``Contour tracking by stochastic propagation of conditional
  density,''
\newblock in {\em Proc. Euro. Conf. Computer Vision}, 1996, pp. 343--356.

\end{thebibliography}
}

\end{document}